\documentclass[letterpaper]{article} 
\usepackage{aaai24}  
\usepackage{times}  
\usepackage{helvet}  
\usepackage{courier}  
\usepackage[hyphens]{url}  
\usepackage{graphicx} 
\usepackage{natbib}  
\usepackage{caption} 
\usepackage{algorithm}
\usepackage{algorithmic}
\usepackage{xcolor}         

\usepackage{graphicx}
\usepackage{xr}
\usepackage{subfigure}
\usepackage{paralist}
\usepackage{amssymb}
\usepackage{amsmath}

\usepackage{newfloat}
\usepackage{listings}
\DeclareCaptionStyle{ruled}{labelfont=normalfont,labelsep=colon,strut=off} 
\floatstyle{ruled}
\newfloat{listing}{tb}{lst}{}
\floatname{listing}{Listing}
\title{Decoupling Meta-Reinforcement Learning with Gaussian Task Contexts and Skills}

\author {
    Hongcai He\textsuperscript{\rm 1},
    Anjie Zhu\textsuperscript{\rm 1},
    Shuang Liang\textsuperscript{\rm 1},
    Feiyu Chen\textsuperscript{\rm 1,2},
    Jie Shao\textsuperscript{\rm 1,2}\thanks{Corresponding author: Jie Shao. This work was supported by the National Natural Science Foundation
of China (grants No. 62276047 and No. 62302080).} }
\affiliations {
    \textsuperscript{\rm 1}University of Electronic Science and Technology of China, Chengdu, China\\
    \textsuperscript{\rm 2}Sichuan Artificial Intelligence Research Institute, Yibin, China\\
    \{hehongcai,anjiezhu\}@std.uestc.edu.cn, \{shuangliang,chenfeiyu,shaojie\}@uestc.edu.cn
}

\usepackage{bibentry}

\begin{document}

\maketitle

\begin{abstract}
Offline meta-reinforcement learning (meta-RL) methods, which adapt
to unseen target tasks with prior experience, are essential in robot
control tasks. Current methods typically utilize task contexts and
skills as prior experience, where task contexts are related to the
information within each task and skills represent a set of
temporally extended actions for solving subtasks. However, these
methods still suffer from limited performance when adapting to
unseen target tasks, mainly because the learned prior experience
lacks generalization, i.e., they are unable to extract effective
prior experience from meta-training tasks by exploration and
learning of continuous latent spaces. We propose a framework called
decoupled meta-reinforcement learning (DCMRL), which (1)
contrastively restricts the learning of task contexts through
pulling in similar task contexts within the same task and pushing
away different task contexts of different tasks, and (2) utilizes a
Gaussian quantization variational autoencoder (GQ-VAE) for
clustering the Gaussian distributions of the task contexts and
skills respectively, and decoupling the exploration and learning
processes of their spaces. These cluster centers which serve as
representative and discrete distributions of task context and skill
are stored in task context codebook and skill codebook,
respectively. DCMRL can acquire generalizable prior experience and
achieve effective adaptation to unseen target tasks during the
meta-testing phase. Experiments in the navigation and robot
manipulation continuous control tasks show that DCMRL is more
effective than previous meta-RL methods with more generalizable
prior experience.
\end{abstract}

\section{Introduction}

Current offline meta-reinforcement learning (meta-RL) methods have
been widely adopted across various domains and produced notable
results, particularly regarding the robot control task
\cite{DBLP:conf/iclr/NamSPHL22, DBLP:conf/icml/RakellyZFLQ19}. These
meta-RL methods acquire prior experience from a series of tasks
during the meta-training phase and then employ the prior experience
to the unseen target tasks which have implicit relationships with
the training tasks during the meta-testing phase. There are two
frequently utilized forms of prior experience, task contexts and
skills \cite{DBLP:conf/iclr/NamSPHL22, DBLP:conf/icml/RakellyZFLQ19,
DBLP:conf/corl/PertschLL20}. Task contexts are related to the vital
statistical information of tasks, which are obtained from past
trajectories generated by agents. Additionally, when meeting an
unseen target task, task contexts that are extracted from its
trajectories will enable agents to acquire its information and
achieve adaptation to the unseen target task
\cite{DBLP:conf/iclr/NamSPHL22, DBLP:conf/icml/RakellyZFLQ19}. On
the other hand, skills represent the means of useful behaviors that
can solve subtasks. As temporal behaviors, skills can be learned
from various forms of data and can be transferred to new tasks and
even new environment configurations. Moreover, a series of skills
for solving different subtasks can be combined to achieve solutions
to complex tasks \cite{DBLP:conf/iclr/NamSPHL22,
DBLP:conf/corl/PertschLL20}.

However, most offline meta-RL methods suffer from poor
generalization issues, hindering them to achieve robust adaptation
to unseen target tasks. This is due to the limited prior experience,
which is caused by the coupled exploration and learning processes of
continuous latent space. More specifically, exploration and learning
processes are interconnected for extracting prior experience from
continuous latent space. Insufficient exploration in the initial
stage often leads to limited learning, which in turn results in
inadequate exploration in subsequent stages. This ultimately results
in both the exploration and learning processes being limited to a
small portion of the entire continuous latent space of prior
experience, leading to sub-optimal decisions
\cite{DBLP:conf/icml/CamposTXSGT20}.

Moreover, the existing methods
\cite{DBLP:conf/icml/ChebotarHLXKVIE21,
DBLP:conf/corl/LynchKXKTLS19, DBLP:conf/corl/PertschLL20,
DBLP:conf/icml/PongNSHL22, DBLP:conf/iclr/NamSPHL22} suffer from
another limitation as they just model task contexts and/or skills as
continuous latent spaces without considering their inherent
characteristics. A task corresponds to a series of similar task
contexts, since agents usually generate diverse trajectories
resulting from variations in their execution processes and levels of
success. For example, in the maze navigation tasks, agents can start
from a fixed point and take different paths to the specified
endpoint. However, failing to consider the relationships between
task contexts of the same and different tasks will result in unclear
and ambiguous learning. In addition, there are a series of similar
skills in the continuous latent space due to the similarity between
subtasks. For example, in the kitchen manipulation tasks, opening
the door of the microwave and opening the hinge cabinet are similar.
Nevertheless, without consideration for relationships between
skills, it will be hard to select the most accurate skill.

To this end, we propose a framework called decoupled
meta-reinforcement learning (DCMRL) using both task contexts and
skills as prior experience so as to acquire task contexts and skills
with generalization, additionally achieving effective adaptation to
unseen target tasks. Specifically, we model the distributions of
task context and skill as Gaussian distributions, instead of just
representing task contexts and skills as simple vectors, since
Gaussian distributions can capture the uncertainty in their
respective spaces and provide more robust representations. Firstly,
DCMRL utilizes our proposed Gaussian quantization variational
autoencoder (GQ-VAE) to perform online clustering on the Gaussian
distributions of task contexts and skills in their respective
continuous latent spaces, generating a set of discrete cluster
centers. These cluster centers are stored in the form of learnable
codebooks, as the representative distributions of task contexts and
skills with generalization. Additionally, exploration of the
continuous latent spaces and learning of discrete cluster centers
inside the codebooks achieve decoupling of exploration and learning
processes, therefore we apply this decoupling operation to both task
contexts and skills via GQ-VAEs. Moreover, task contexts are
vulnerable to the distribution mismatch of meta-training tasks and
unseen target tasks during meta-testing. To solve this issue, DCMRL
contrastively restricts task contexts through the dissimilarity of
task contexts for different tasks and the similarity of different
task contexts for the same task, leading to task contexts with
generalization. In essence, DCMRL enhances the generalizability of
the task contexts and skills acquired as prior experience during the
meta-training phase, thereby enabling more effective adaptation to
unseen target tasks during the meta-testing phase.

The main contributions of our method are threefold:
\begin{compactitem}
\item We propose DCMRL, a novel framework that enhances the
generalizability of task contexts and skills by contrastively
restricting task contexts and decoupling the exploration and
learning of their respective spaces, leading to more effective
adaptation to unseen target tasks.
\item We propose a novel GQ-VAE that clusters on Gaussian distributions of
task context and skill distributions in their corresponding
continuous latent spaces and decouples the exploration and learning
of their respective spaces, enhancing their generalizability.
\item We evaluate DCMRL in two challenging continuous robot control
environments, i.e., maze navigation and kitchen manipulation, which are
long-horizon and sparse-reward. The results show that DCMRL
outperforms previous meta-RL methods, achieving
more effective adaptation to unseen target tasks.
\end{compactitem}

\section{Related Work}

\paragraph{Offline Meta-reinforcement Learning.}

The primary goal of offline meta-reinforcement learning (meta-RL) is
the acquisition of learning strategies from offline datasets,
allowing more effective learning in new tasks through appropriate
prior experience. Due to a distributional shift between offline and
online data during testing, it is critical to obtain robust task
representations that generalize well while learning. Most previous
methods \cite{DBLP:conf/nips/DorfmanST21,
DBLP:conf/icml/MitchellRPLF21, DBLP:conf/icml/PongNSHL22,
DBLP:conf/iclr/SiegelSBANLHHR20} meta-learn from offline datasets,
including reward and task annotations, and adapt to a new task with
only a small amount of new data. However, some meta-training tasks
are hard to annotate due to the lack of corresponding prior task
experience. Moreover, \citet{DBLP:conf/icml/PongNSHL22} utilize
semi-supervised learning with both offline and online data for
distributional shift, but heavily relying on annotation functions
from offline data. In contrast, our proposed DCMRL leverages a large
offline dataset across many tasks without rewards or task
annotations for extracting skills.

\paragraph{Context-based Meta-RL.}

Context-based methods train a module to take prior experience as a
form of task-specific context. Some methods
\cite{DBLP:journals/corr/DuanSCBSA16, DBLP:conf/icml/FinnAL17,
DBLP:journals/corr/abs-1905-06424, DBLP:conf/icml/LiuRLF21,
DBLP:conf/iclr/RothfussLCAA19, DBLP:conf/cogsci/WangKSLTMBKB17,
DBLP:conf/iclr/YuFXDZAL18, DBLP:conf/atal/YangCITF19,
DBLP:conf/icml/ZintgrafSKHW19} have been proposed for meta-learning
dynamic models and policies that can quickly adapt to unseen target
tasks. In contrast, other recursive methods
\cite{DBLP:conf/iclr/FakoorCSS20, DBLP:conf/icml/LeeSLLS20,
DBLP:conf/iclr/MishraR0A18, DBLP:conf/icml/RakellyZFLQ19,
DBLP:conf/nips/SeoLGKSA20} make fast adaptation by aggregating
experience into a latent representation on which the policy is
conditioned. Additionally, some methods train recurrent Q-function
with off-policy Q-learning approaches which are often used on simple
tasks \cite{DBLP:journals/corr/HeessHLS15} or in discrete
environments \cite{DBLP:conf/aaaifs/HausknechtS15}. As a
context-based method, DCMRL decouples the exploration and learning
processes of task contexts during meta-training for improving the
generalization of task contexts, and achieves effective adaptation
to unseen target tasks during meta-testing.

\paragraph{Skill-based Meta-RL.}

Another method for exploiting offline data without requiring reward
or task annotations is extracting skills as the identification of
reusable, short-horizon behaviors. Skill-based learning methods
learn unseen long-horizon target tasks by transferring these learned
skills and converge significantly faster than learning from scratch
\cite{DBLP:conf/iclr/HausmanS0HR18, DBLP:conf/iclr/LeeSSHL19}.
Previous works \cite{DBLP:conf/iclr/AjayKALN21,
DBLP:conf/icml/ChebotarHLXKVIE21, DBLP:conf/corl/LynchKXKTLS19,
DBLP:journals/tog/MerelTATHPEWH20, DBLP:conf/corl/PertschLL20,
DBLP:conf/corl/PertschLWL21} have shown that skill-based learning
methods can learn a broad range of skills with diverse datasets and
accomplish long-horizon tasks. However, these methods still require
a substantial number of interactions with environment to learn
enough skills or new skills. SiMPL \cite{DBLP:conf/iclr/NamSPHL22}
learns skills by combining meta-learning process and offline dataset
but still suffers from limited generalization. As a skill-based
method, DCMRL further applies the decoupling operation to the
exploration and learning processes of skills, generating more
generalized and representative skills.

\begin{figure*}[t]
  \centering
  \includegraphics[width=0.7\textwidth]{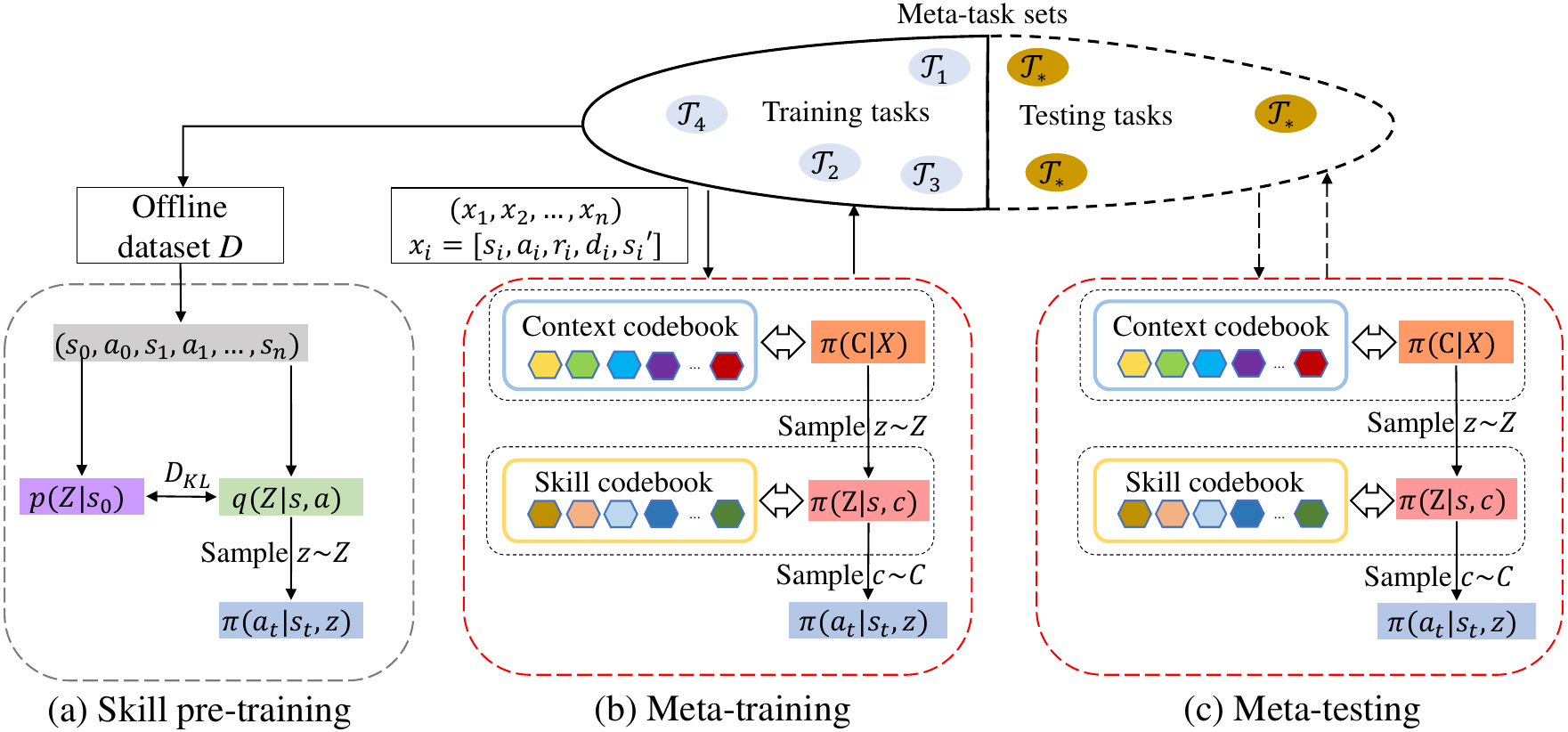}
  \caption{\textbf{Method overview.} DCMRL as a hierarchical framework, has
three phases. \textbf{(a) Skill pre-training} learns a prior skill
$p(Z|s)$ (\textcolor[RGB]{204,153,255}{purple}) as a constraint and
a low-level skill-based policy $\pi(a_t|s_t,z)$
(\textcolor[RGB]{180,199,231}{blue}) trained with $q(Z|s,a)$
(\textcolor[RGB]{197,224,180}{green}) through the offline dataset
$D$. \textbf{(b) Meta-training} meta-trains a high-level skill
policy $\pi(Z|s,c)$ (\textcolor[RGB]{255,153,153}{red}) for skill
distribution $Z$ and a task context policy $\pi(C|\mathit{X})$
(\textcolor[RGB]{255,153,102}{orange}) for task context
distributions $C$ through meta-training tasks
$T=\{\mathcal{T}_1,...,\mathcal{T}_{n_{task}}\}$, while the
pre-trained low-level skill-based policy $\pi(a_t|s_t,z)$ remains
fixed. \textbf{(c) Meta-testing} utilizes the meta-trained modules
for effective adaptation to an unseen target task $\mathcal{T}^*\in
T^*$ with task contexts generated from a few transitions of it.
Additionally, $\pi(C|\mathit{X})$ and $\pi(a_t|s_t,z)$ are fixed and
$\pi(Z|s,c)$ remains under fine-tuning. } \label{fig:structure}
\end{figure*}

\section{Method}
\label{sec:method}

Decoupled meta-reinforcement learning (DCMRL) consists of three
phases: skill pre-training, meta-training and meta-testing. We
mainly focus on the meta-training phase and aim to acquire more
generalizable task contexts and skills through: (1) contrastively
restricting task contexts by the relationships between tasks for
enhancing their generalization and (2) applying our proposed GQ-VAEs
to cluster the distributions of task contexts and skills
respectively, and decouple the exploration and learning processes of
their respective spaces. An illustration of DCMRL is given in
Figure~\ref{fig:structure}.

\subsection{Problem Formulation}

An offline meta-RL task is typically formalized as a fully
observable Markov decision process (MDP), defined as a tuple
$<\mathcal{S}, \mathcal{A}, p, r, \gamma, \rho_0>$. Here,
$\mathcal{S}$ is the state space, $\mathcal{A}$ is the action space,
$p(s'|s, a)$ is the transition dynamics, $r(s, a)$ is the reward
function, $\rho_0$ is the initial state distribution, and $\gamma
\in [0, 1)$ is the factor discounting the future reward. The policy
is a distribution $\pi(a|s)$ over actions. In a complete MDP, the
agent is initialized in a given state and selects an action at each
time step by sampling from a fixed policy $\pi$. Meanwhile, the
environment responds by updating the state using transition
probabilities $p$ and providing a reward $r$ and a boolean flag of
done $d$. Additionally, the marginal state distribution at time step
$t$ is defined as $\mu_\pi^t(s)$ and the objective of the agent is
to maximize the expected cumulative rewards
$max_\pi\mathcal{J}_\mathcal{M}(\pi) =
\mathbb{E}_{s_t\sim\mu_\pi^t,a_t\sim\pi}[\sum^\infty_{t=0}\gamma^t
r(s_t, a_t)]$. More details of problem formulation and preliminaries
can be found in Appendix.

As an offline meta-RL method, DCMRL assumes access to a
\textit{task-agnostic} dataset of state-action trajectories $D =
(s_0, a_0, \ldots, s_t, a_t)$, collected from various tasks or
unlabeled data. With a wide variety of behaviors, $D$ is able to
accelerate learning of different tasks
\cite{DBLP:conf/iclr/NamSPHL22}. We also assume access to a set of
$n_{task}$ meta-training tasks $T = \{\mathcal{T}_1, \ldots,
\mathcal{T}_{n_{task}} \}$, simultaneously with a set of unseen
target tasks $T^*$, and represent each task as an MDP respectively.
Notably, we do not assume that there are direct relationships
between either $T^*$ and $D$ or $T^*$ and $T$. Specifically, the
offline dataset $D$ does not contain any demonstrations for solving
tasks in $T^*$, and there are no intersections between $T^*$ and
$T$.

DCMRL aims to extract skills from the offline dataset $D$, perform
the meta-training phase on the set of meta-training tasks $T$ and
handle the set of unseen target tasks $T^*$ in the meta-testing
phase. Moreover, we represent the distributions of task context and
skill as Gaussian distributions, denoted by $C$ and $Z$, while $c$
and $z$ are the task context embeddings and skill embeddings sampled
from them.

\subsection{Skill Pre-training}

During the skill pre-training phase, DCMRL comprises a skill prior
$p(Z|s_0)$, a skill encoder $q(Z|s,a)$ and a skill-based policy
$\pi(a_t|s_t,z)$. Specifically, our primary focus lies on the skill
prior $p(Z|s_0)$ and skill-based policy $\pi(a_t|s_t,z)$. Both the
skill prior $p(Z|s_0)$ and skill encoder $q(Z|s,a)$ are implemented
as deep neural networks that output skill distributions $Z$ in the
form of Gaussian distributions. For a K-step trajectory randomly
drawn from the sequences in offline dataset $D$, the skill prior
$p(Z|s_0)$ predicts a skill distribution $Z$ based on the initial
state $s_0$ of the trajectory, while the skill encoder $q(Z|s,a)$
aligns the sequence of full state-action pairs to a skill
distribution $Z$. Moreover, the skill prior $p(Z|s_0)$ is trained by
matching to the skill distribution encoded by the skill encoder
$q(Z|s,a)$ as follows:
\begin{equation} \small
\mathop{min}_{p}\mathcal{D}_{KL}(sg[q(Z|s,a)],p(Z|s_0)),
\end{equation}
where $sg[\cdot]$ denotes the stop-gradient operation and
$\mathcal{D}_{KL}$ denotes the Kullback-Leibler divergence.
Furthermore, skill-based policy $\pi(a_t|s_t,z)$ is fine-tuned
through behavioral cloning to replicate the action sequence
$a_{0:K-1}$ corresponding to the given skill embedding $z$ which is
sample from the skill distribution $Z$ output by skill encoder
$q(Z|s,a)$. Moreover, we leverage a unit Gaussian prior distribution
of skill inspired by \citet{DBLP:conf/iclr/HigginsMPBGBML17} for
regularization:
\begin{equation} \small
\begin{aligned}
  \max_{q,\pi}\mathbb{E}_{z \sim q}[&\prod_{t=0}^{K-1} \log\pi(a_t|s_t,z) \\
  &- \alpha \mathcal{D}_{KL}(q(Z|s,a),\mathcal{N}(0,\mathcal{I}))],
\end{aligned}
\end{equation}
where $\alpha$ is a weight coefficient of the constraint. In
summary, we make a skill pre-training phase for skill prior
$p(Z|s_0)$, which is utilized for imposing constraints on high-level
skill policy $\pi(Z|s,c)$, and skill-based policy $\pi(a|s,z)$,
which serves as the low-level policy and remains fixed during
meta-training and meta-testing.

\subsection{Meta-training}

\subsection{Contrastive Task Context Learning}

\begin{figure}[t]
  \centering
  \includegraphics[width=0.5\textwidth]{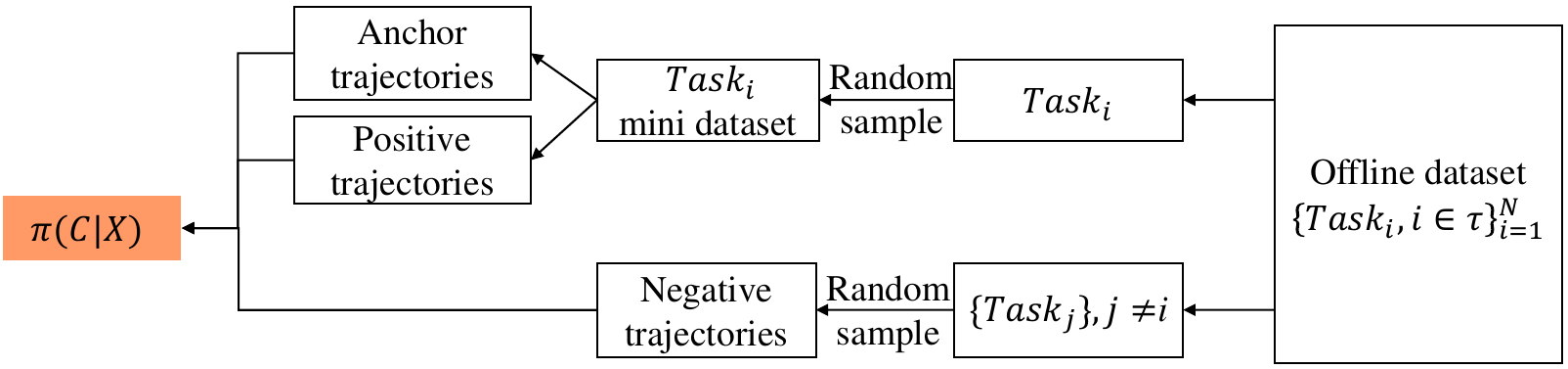}
  \caption{\textbf{Contrastive task context learning.} DCMRL samples a long
trajectory data as the mini dataset and generates anchor
trajectories and positive trajectories from it, while negative
trajectories are from other tasks.}
  \label{fig:constrasive}
\end{figure}

Task context as a kind of prior experience, aims to specify which
task from the distribution the agent should focus on, and provides
information that helps the agent adapt to its strategy when a new
task is encountered. We contrastively enhance task context
representations by distinguishing unique task features, thereby
promoting effective and human-like adaptability in diverse tasks.

We employ a specific sampling strategy on trajectories from
meta-training tasks. Traditional methods such as
\citet{DBLP:conf/icml/YuanL22} sample positive samples from the same
task, but this can result in chaos and significant disparity from
the anchor trajectory due to variations in tuples and their orders.
Hence, DCMRL samples anchor and positive samples in two stages:
first, a longer trajectory is sampled as a mini dataset from current
task; then, two different trajectories of the same and fixed length
are sampled from this long trajectory, maintaining tuples' relative
orders, to serve as the anchor and positive samples respectively.
Negative samples are randomly sourced from other tasks, as
traditional methods (see Figure~\ref{fig:constrasive}).

We utilize the classic contrastive learning loss function, triplet
loss \cite{DBLP:conf/cvpr/SchroffKP15}, for anchor, positive, and
negative samples. These trajectories are processed with a high-level
task context policy $\pi(C|X)$ to generate distributions of task
contexts, $\tilde{C}$, $\tilde{C}^+$, and $\tilde{C}^-$. The triplet
loss aims to minimize the similarity between $\tilde{C}$ and
$\tilde{C}^-$ and maximize the similarity between $\tilde{C}$ and
$\tilde{C}^+$ as follows:
\begin{equation} \small
\begin{aligned}
  \mathcal{L}_{triplet}(\tilde{C}, \tilde{C}^+, \tilde{C}^-)=\sum_{\tilde{C}\in \mathcal{C}}&max(0,sim(\tilde{C}, \tilde{C}^-) \\
  &-sim(\tilde{C},\tilde{C}^+)+\epsilon),
\end{aligned}
\end{equation}
where $sim(\cdot)$ is the similarity function, and we utilize the
cosine similarity here. $\mathcal{C}$ is the continuous latent space
of task contexts and the margin parameter $\epsilon$ is configured
as the minimum offset between distances of similar and dissimilar
pairs.

\begin{figure*}[t]
  \centering
  \includegraphics[width=0.72\textwidth]{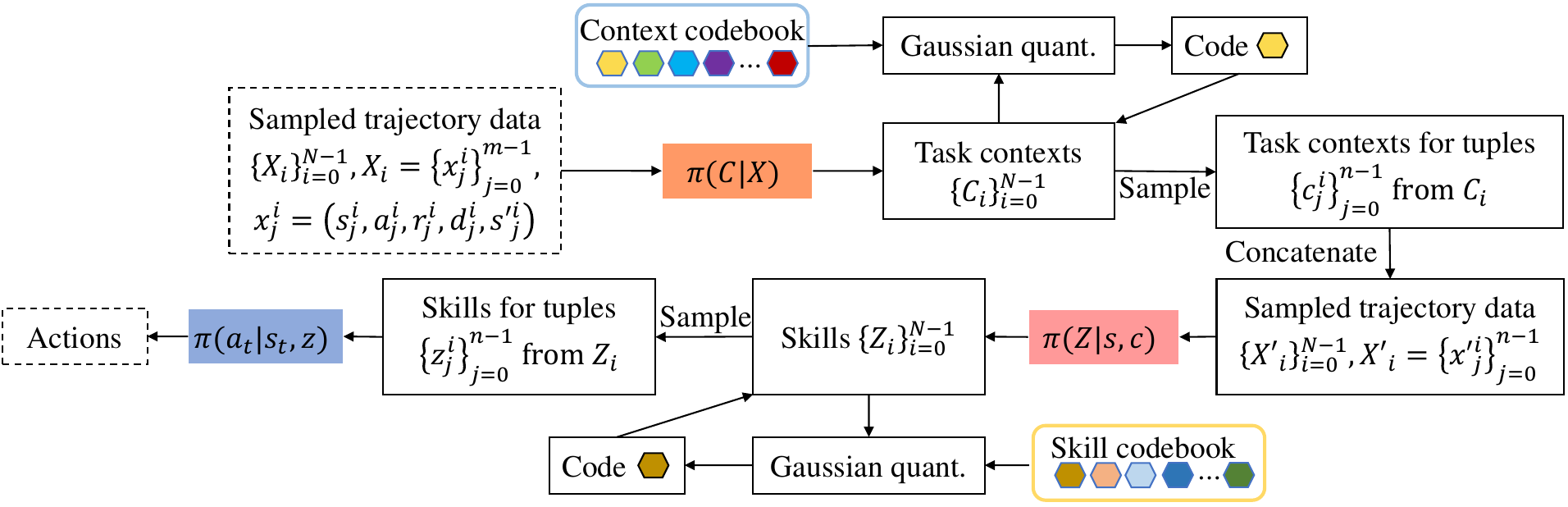}
  \caption{\textbf{Task context and skill learning architecture.} The task
context distribution $C$ and skill distribution $Z$ generated by
task policy $\pi(C|\mathit{X})$ and skill policy $\pi(Z|s,c)$
respectively, are both implemented in the form of Gaussian
distributions. The task context embeddings $c$ and skill embeddings
$z$ are sampled from these distributions. In addition, the learning
processes of $\pi(C|\mathit{X})$ and $\pi(Z|s,c)$ are accompanied by
corresponding learnable codebooks that correspond to their
respective spaces.}
  \label{fig:VQ}
\end{figure*}

\subsection{GQ-VAE}

We propose Gaussian quantization variational autoencoder (GQ-VAE),
which consists of three main parts: an encoder, a learnable codebook
and a decoder, where the encoder and decoder are deep neural
networks. The codebook $\mathcal{CB} = \{O^1, \ldots, O^K\}$
contains $K$ cluster centers as discrete codes, where $K$ is a
hyperparameter.

Specifically, the encoder maps the input trajectory $\mathit{X}$ to
a Gaussian distribution $\tilde{O}$. Codes within $\mathcal{CB}$ are
modeled as Gaussian distributions, as the targets for clustering
originate from the Gaussian distributions encoded by the encoder.
Once a Gaussian distribution $\tilde{O}$ is outputted by the
encoder, it will be matched to its closest code $O^k$ within
$\mathcal{CB}$ through a match operation $\mathbf{m}(\cdot)$.
Moreover, Euclidean distance is utilized to measure the distance
between current $\tilde{O}$ and each $O^i$, where $1\le i \le K$ in
the codebook $\mathcal{CB}$. The complete process of
$\mathbf{m}(\cdot)$ is as follows:
\begin{equation} \small
  O=\mathbf{m}(\tilde{O}):= \mathop{argmin}_{O^k \in \mathcal{CB}, 1 \le k \le K}\Vert\tilde{O}-O^k\Vert_2.
\end{equation}
Finally, the decoder will take the code $O^k$ that matches
$\tilde{O}$ as input to output a reconstructed trajectory
$\tilde{\mathit{X}}$.

In essence, exploration of the latent space and learning of codebook
$\mathcal{CB}$ in DCMRL interacts to yield complementary and
reinforcing effects. Directly, exploring the continuous latent space
forms its composition in Gaussian distributions, while learning in
codebook $\mathcal{CB}$ achieves discretization of the continuous
latent space by deriving $K$ codes as cluster centers. Moreover, the
match operation $\mathbf{m}(\cdot)$ integrates the two stages by
matching the initialized codes in the codebook $\mathcal{CB}$ with
different $\tilde{O}$. The codes learn from different $\tilde{O}$
they match and optimize their positions in the continuous latent
space. Overall, this online clustering procedure resembles a
classical $K$-means algorithm. The loss function used to update
GQ-VAE is as follows:
\begin{equation} \small
  \mathcal{L}_{GQ} = \Vert sg[\mathbf{m}(\tilde{O})]-\tilde{O}\Vert _2 + \mu\Vert \mathbf{m}(\tilde{O})-sg[\tilde{O}]\Vert _2 + \Vert \tilde{\mathit{X}}-\mathit{X}\Vert _2,
\end{equation}
where $\mu$ is a weight coefficient.

The loss function comprises three terms. The first two terms differ
solely in the objects of the $sg[\cdot]$ operation, updating the
encoder and matched code in $\mathcal{CB}$, respectively, and $\mu$
balances them. The third term is derived by inputting the matched
code after the $sg[\cdot]$ operation to the decoder, which optimizes
the decoder through quantifying the disparity between reconstructed
and input trajectories.

In summary, GQ-VAE decouples the exploration over the continuous
latent space and the learning of the codebook $\mathcal{CB}$. Its
bidirectional updating mechanism achieves better learning of each
code within $\mathcal{CB}$ through exploration, facilitating
discretization. Meanwhile, exploration guided by learned codes
attains deeper characterization. DCMRL simultaneously utilizes task
contexts and skills as prior experience, and applies GQ-VAEs on both
task contexts and skills as shown in Figure~\ref{fig:VQ}. Next, we
introduce the exploration of the two continuous latent spaces of
task contexts and skills, as well as the learning of corresponding
task context codebook and skill codebook.

\paragraph{Task Context Learning.}

GQ-VAE for the task context learning stage includes three modules: a
high-level task context policy $\pi(C|\mathit{X})$ as task context
encoder, a learnable codebook of task context with $K_\mathcal{C}$
quantized task context codes $\mathcal{CB_C}=\{C^1, ...,
C^{K_\mathcal{C}}\}$ and a corresponding decoder.

Given a batch of sampled trajectory data $\mathcal{X} =
\{\mathit{X}_1, \dots, \mathit{X}_N\}$, where $N$ is the task batch
size that means the sampled number of tasks and $X_i = \{(s^{i}_{j},
a^{i}_{j}, r^{i}_{j}, d^{i}_{j}, s'^{i}_{j})\}_{j=0}^{n_c}$ is input
task trajectory data whose length is $n_c$, the task context encoder
$\pi(C|\mathit{X})$ outputs a task context distribution $\tilde{C}$.
The complete process of matching is as follows:
\begin{equation} \small
  C=\mathbf{m}(\tilde{C}):= \mathop{argmin}_{C^k \in \mathcal{CB_C}, 1 \le k \le K_\mathcal{C}}\Vert\tilde{C}-C^k\Vert_2.
\end{equation}
We utilize an objective $\mathcal{L}_{Context} = \mathcal{L}_{BC} +
\lambda\mathcal{L}_{GQ_{Context}}$ for updating, where
$\mathcal{L}_{BC}$ is the behavior-cloning loss generated by the
skill-based policy $\pi(a_t|s_t,z)$, $\lambda$ is a weight
coefficient of loss and the formulation of
$\mathcal{L}_{GQ_{Context}}$ is as follows:
\begin{equation}
\begin{aligned} \small
  \mathcal{L}_{GQ_{Context}} = \Vert sg[\mathbf{m}(\tilde{C})]-\tilde{C}\Vert _2 &+ \eta\Vert \mathbf{m}(\tilde{C})-sg[\tilde{C}]\Vert _2 \\
  &+ \Vert \tilde{\mathit{X}}-\mathit{X}\Vert _2,
\end{aligned}
\end{equation}
where $\eta$ is a weight coefficient.

\paragraph{Skill Learning.}

GQ-VAE for the skill context learning stage includes a high-level
skill policy $\pi(Z|s,c)$ as skill encoder, a learnable codebook of
skills with $K_\mathcal{Z}$ quantized skill codes $\mathcal{CB_Z} =
\{Z^1, \ldots, Z^{K_\mathcal{Z}}\}$ and a decoder.

Given an additional batch of sampled trajectories $\mathcal{X}' =
\{\mathit{X}_1', \ldots, \mathit{X}_N'\}$ from the same tasks, it is
different from that used in the task context stage and the length of
sampled trajectories $n_z$ is often different from $n_c$. The skill
policy $\pi(Z|s,c)$ inputs a sampled trajectory $\mathit{X}_i'$ and
task context $c$ sampled from $C$ to output a skill distribution
$\tilde{Z}$. Furthermore, the complete process of matching is as
follows:
\begin{equation} \small
  Z=\mathbf{m}(\tilde{Z}):= \mathop{argmin}_{Z^k \in \mathcal{CB_Z}, 1 \le k \le K_\mathcal{Z}}\Vert\tilde{Z}-Z^k\Vert_2.
\end{equation}
The objective is $\mathcal{L}_{Skill} = \mathcal{L}_{BC} +
\gamma\mathcal{L}_{GQ_{Skill}}$, where $\gamma$ is a weight
coefficient and $\mathcal{L}_{GQ_{Skill}}$ is the skill quantization
loss:
\begin{equation}
\begin{aligned} \small
  \mathcal{L}_{GQ_{Skill}} = \Vert sg[\mathbf{m}(\tilde{Z})]-\tilde{Z}\Vert _2 &+ \iota\Vert \mathbf{m}(\tilde{Z})-sg[\tilde{Z}]\Vert _2 \\
  &+ \Vert \tilde{\mathit{X'}}-\mathit{X'}\Vert _2,
\label{Skill_GQ-VAE_loss}
\end{aligned}
\end{equation}
where $\iota$ is also a weight coefficient.

Besides applying GQ-VAE, the skill policy updating also leverages
the skill prior $p(Z|s_0)$ from the skill pre-training phase, as
\citet{DBLP:conf/iclr/NamSPHL22}, with the objective as follows:
\begin{equation} \small
\begin{aligned}
\mathop{\max}_\pi &\mathbb{E}_{c \sim \pi(\cdot|X)}[\sum_t \mathbb{E}_{(s_t, z) \sim \rho_{\pi|c}}[ r_{\mathcal{T}}(s_t, z) \\
& - \beta \mathcal{D}_{KL}(\pi(Z|s,c),p(Z|s_0))]],
\label{Skill_prior_loss}
\end{aligned}
\end{equation}
where $\beta$ is a weight coefficient of the constraint.

In the subsequent process, skill-based policy $\pi(a_t|s_t,z)$ uses
specific skill $z$, which is sampled from $Z$ and current state, to
general corresponding action.

\subsection{Meta-testing}

DCMRL has trained the high-level task context policy
$\pi(C|\mathit{X})$ and skill policy $\pi(Z|s,c)$ on the set of
meta-training tasks during meta-training phase. When facing unseen
target tasks in the meta-testing phase, we first collect a few
trajectories $\mathit{X}^*$ and extract the task context
distribution $C^*$ from them through the task context policy
$\pi(C|\mathit{X})$. Then, we utilize the skill policy $\pi(Z|s,c)$
under $c^*$ sampled from $C^*$ to generate $Z^*$ and sample $z^*$
from it. In order to refine DCMRL on the unseen target tasks, we
continue to optimize the skill policy through
Eq.~\eqref{Skill_GQ-VAE_loss} and Eq.~\eqref{Skill_prior_loss}.

\begin{figure}[t]
\centering
\includegraphics[width=0.45\linewidth]{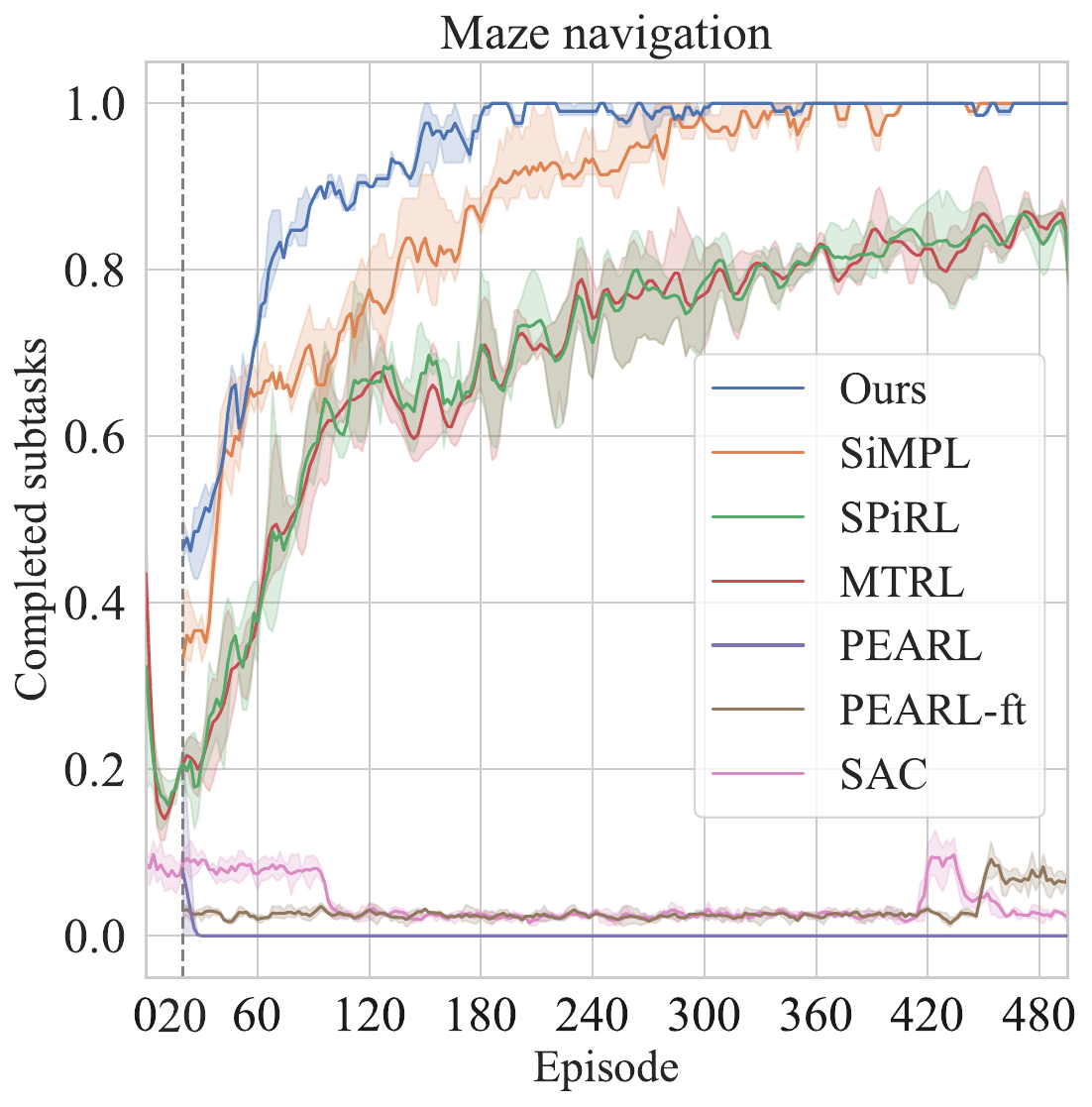}
\includegraphics[width=0.45\linewidth]{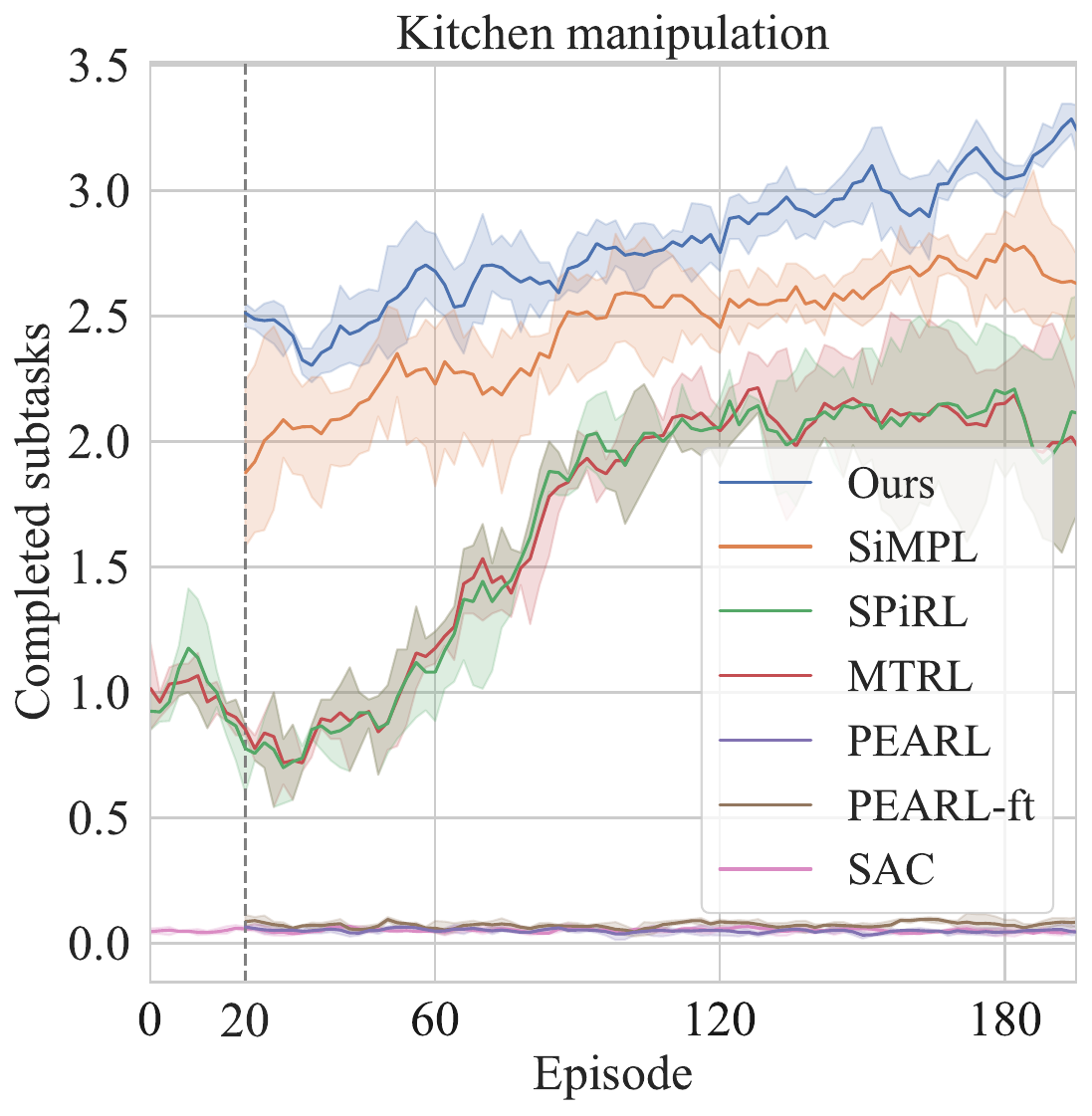}
\caption{\textbf{Comparisons of sample efficiency.} We evaluate
DCMRL, SiMPL, SPiRL and MTRL in maze navigation and kitchen
manipulation. In both environments, we individually train each model
for every target task using five distinct random seeds. Prior to
fine-tuning on the target tasks, our method, SiMPL, PEARL and
PEARL-ft employ 20 episodes of environment interactions for
conditioning the meta-trained policies.} \label{fig:comparison}
\end{figure}

\begin{figure}[t]
\centering
\includegraphics[width=0.6\linewidth]{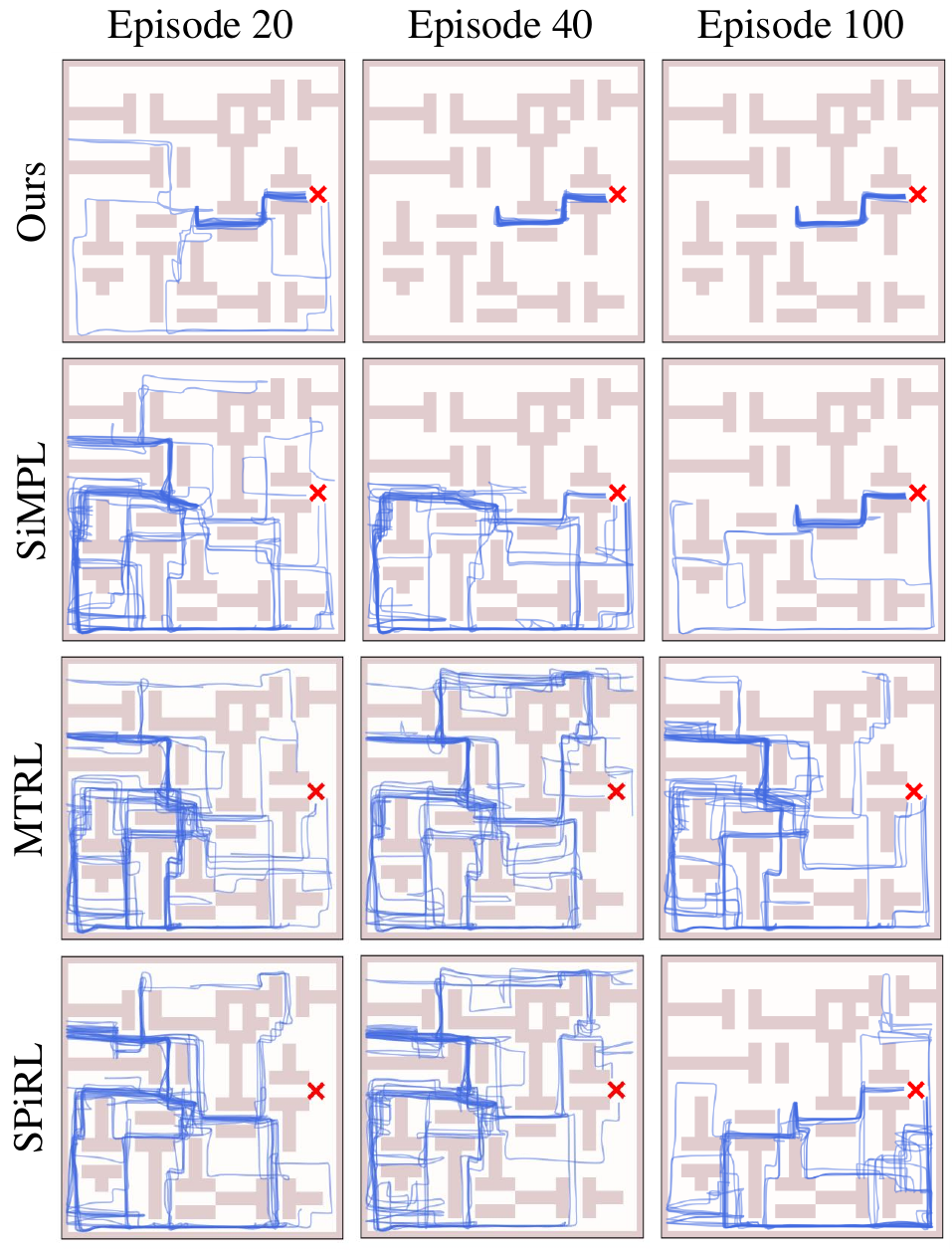}
\caption{\textbf{Visualization of the adaptation process.} DCMRL,
SiMPL, SPiRL and MTRL can make effective adaptation of the maze
navigation environment with prior experience. We present their
adaptation situations in episodes 20, 40 and 100.}
\label{fig:Maze_environment_compare}
\end{figure}

\section{Experiments}

Our experiments are mainly based on long-horizon and sparse-reward
tasks and evaluate DCMRL on two key issues: (1) whether better prior
experience can be learned and (2) whether the effective adaptation
of unseen target tasks can be achieved. Our code is available at
\url{https://github.com/hehongc/DCMRL/}.

\begin{figure*}
\centering
\subfigure[Sparser task distribution]{
    \includegraphics[width=0.28\textwidth]{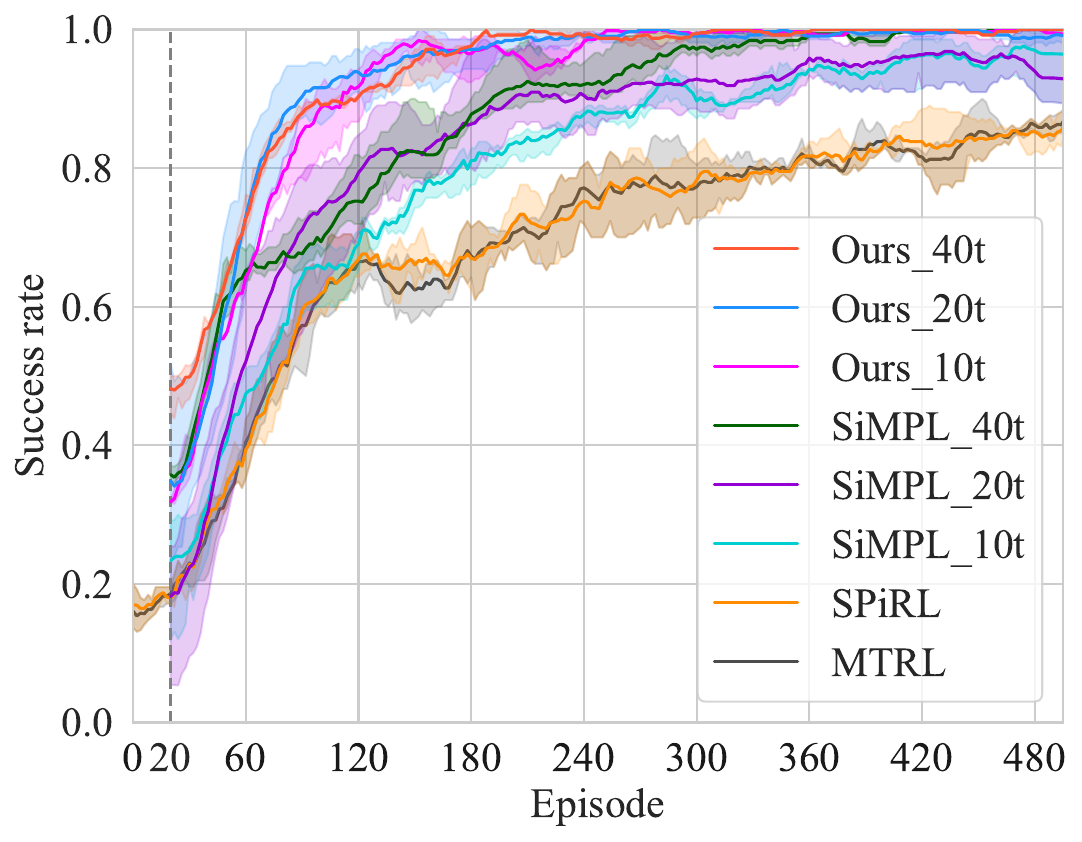}
}
\subfigure[$\mathcal{T}_{Train-top}\to\mathcal{T}_{Target-top}$]{
    \includegraphics[width=0.28\textwidth]{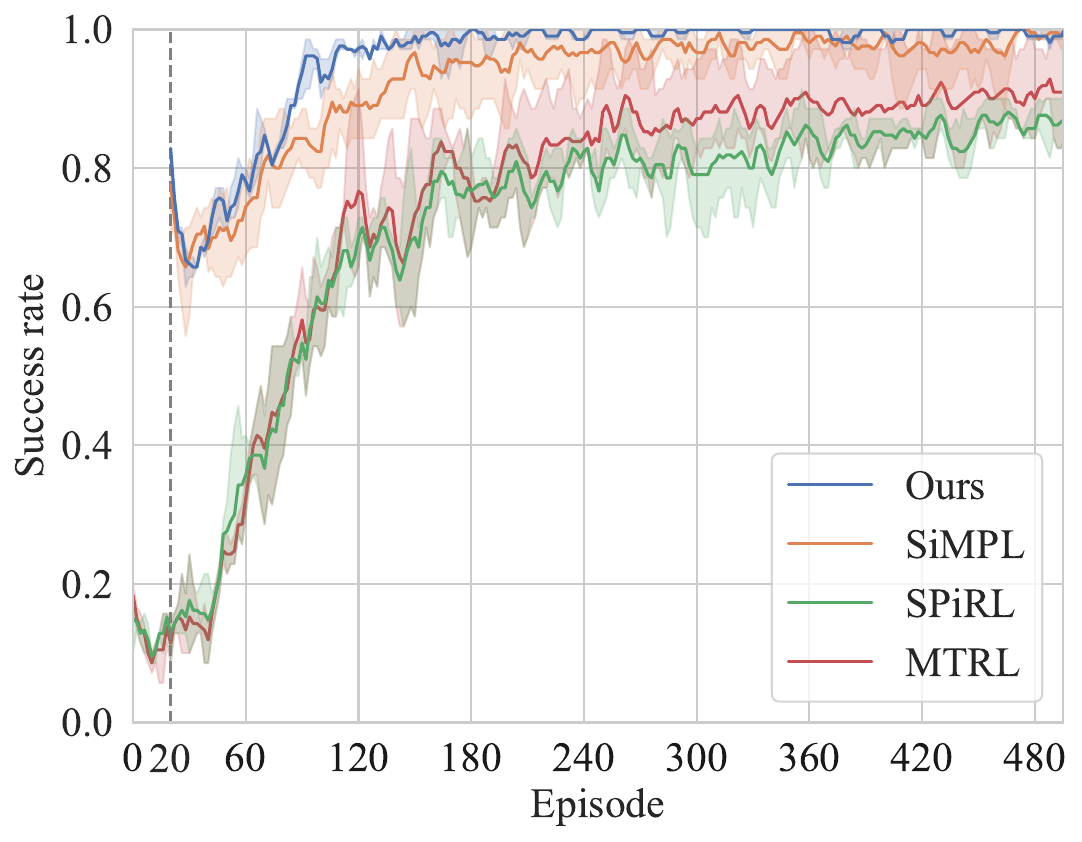}
}
\subfigure[$\mathcal{T}_{Train-top}\to\mathcal{T}_{Target-bottom}$]{
    \includegraphics[width=0.28\textwidth]{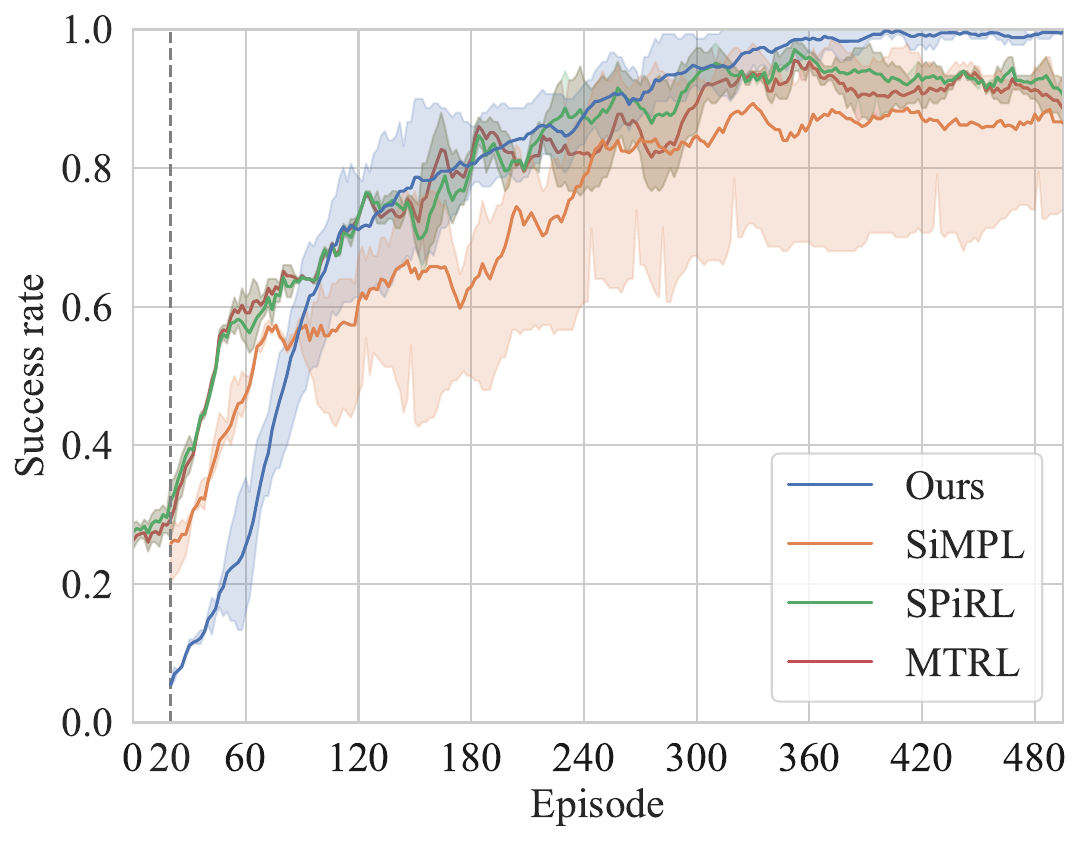}
} \caption{\textbf{Meta-training and target task distribution
analysis.} (a) DCMRL and SiMPL are trained on both the original
meta-training task number setup (40) and lower setup (i.e., 10, 20),
denoted as 10t, 20t and 40t, while SPiRL and MTRL are only trained
on the original setup. All models are evaluated with the same set of
unseen target tasks. (b) DCMRL, SiMPL, SPiRL and MTRL are trained on
a meta-training task distribution ($\mathcal{T}_{Train-top}$) that
exhibits higher alignment with the target task distribution
($\mathcal{T}_{Target-top}$). (c) DCMRL, SiMPL, SPiRL and MTRL are
trained on a meta-training task distribution
($\mathcal{T}_{Train-top}$) which is misaligned with the target task
distribution ($\mathcal{T}_{Target-bottom}$). To comprehensively
evaluate the efficacy of our approach, we train each model on each
target task using five distinct random seeds.}
\label{fig:more_comparison_Maze}
\end{figure*}

\subsection{Experimental Setup}

We compare DCMRL with SiMPL \cite{DBLP:conf/iclr/NamSPHL22}, SAC
\cite{DBLP:conf/icml/HaarnojaZAL18}, SPiRL
\cite{DBLP:conf/corl/PertschLL20}, PEARL
\cite{DBLP:conf/icml/RakellyZFLQ19}, PEARL-ft
\cite{DBLP:conf/iclr/NamSPHL22} and MTRL
\cite{DBLP:conf/nips/TehBCQKHHP17}. We evaluate in two complex
continuous control environments: maze navigation and kitchen
manipulation. Maze navigation is a 2D environment, in which the
agent typically requires hundreds of time steps to complete a task,
and only sparse rewards are provided upon success. Kitchen
manipulation involves a 7-DoF robotic arm for executing a task
consisting of four subtasks, in which the agent generally takes
300-500 time steps to complete a task, and only sparse rewards are
provided after complete subtasks in order. More details about the
experimental environments and baselines are in Appendix.

\subsection{Comparison with State-of-the-art Methods}

We report both quantitative performance and qualitative adaptation
experimental results, presented in Figures \ref{fig:comparison} and
\ref{fig:Maze_environment_compare} respectively. Specifically,
Figure~\ref{fig:comparison} shows key insights on adaptation to
unseen target tasks and is used to evaluate the performance of DCMRL
in a quantitative manner. Meanwhile,
Figure~\ref{fig:Maze_environment_compare} offers additional
inspection and verification of DCMRL for an intricate maze
navigation domain in terms of qualitative analysis. DCMRL exhibits
superior performance and sample efficiency compared with all
baselines in Figure~\ref{fig:comparison} for adapting to unseen
target tasks. Additionally, ablation experiments can be found in
Appendix.

We delve into the impact of leveraging prior experience in
reinforcement learning methods to adapt to unseen target tasks.
Without leveraging prior experience, SAC exhibits constrained
adaptation. While PEARL and PEARL-ft learn task contexts as prior
experience from meta-training tasks, they struggle to effectively
adapt to unseen target tasks even with fine-tuning. In addition,
SPiRL leverages a series of continuous skills extracted from offline
datasets as prior experience, which provide limited assistance for
adaptation. Moreover, MTRL trains a multi-task agent from
meta-training tasks, exhibiting adaptation performance comparable to
SPiRL. Furthermore, SiMPL utilizes both skills and task contexts,
which are extracted respectively from offline datasets and
meta-training tasks, as prior experience. However, its adaptation
remains limited due to the constrained generalizability of prior
experience.

Generally, DCMRL exhibits significantly quicker and better unseen
target task adaptation than other methods. In just a few episodes,
it achieves policy convergence to solve nearly 90\% of the unseen
target tasks in the maze environment and nearly three out of four
subtasks in the kitchen manipulation environment. Subsequently, the
further learning of skills for adaptation is able to achieve better
performance. Finally, DCMRL can solve nearly 100\% of the unseen
target tasks in the maze environment and over three out of four
subtasks in the kitchen manipulation environment.

The visual representations depicted in
Figure~\ref{fig:Maze_environment_compare} demonstrate that the
application of offline datasets by DCMRL, SiMPL, SPiRL, and MTRL
lead to effective adaptation of the maze environment in the early
episodes. DCMRL outperforms SiMPL, SPiRL and MTRL in terms of
convergence speed, achieving higher sample efficiency.

\subsection{Meta-training \& Target Task Distribution Analysis}

In this section, we explore how the meta-training task distribution
impacts the adaptation of unseen target tasks. Our evaluation
focuses on two specific factors: (1) the quantity of tasks in the
meta-training task distribution, and (2) the alignment of the
meta-training task distribution with the target task distribution.
Our experiments are conducted within the context of the maze
navigation task.

\paragraph{The Quantity of Meta-training Tasks.}

We aim to gauge the extent to which varying the number of
meta-training tasks influences adaptation. Following
\citet{DBLP:conf/iclr/NamSPHL22}, we train DCMRL with a lower
quantity of meta-training tasks (i.e., 10 and 20) in addition to the
original number setup (40), and evaluate these models with the same
set of unseen target tasks. The quantitative results presented in
Figure~\ref{fig:more_comparison_Maze}(a) indicate that even with
fewer numbers of meta-training tasks, DCMRL exhibits similar
performance and exceeds the performance of best baseline in all
settings (i.e., SiMPL).

\paragraph{Alignment of Meta-training and Target Tasks.}

The focus of this investigation is to determine the extent to which
a model's performance would improve or deteriorate when trained on a
meta-training task distribution that aligns differently with the
target tasks. To achieve this objective, we implement task
distributions that possess varied degrees of bias towards either
meta-training or target tasks. Specifically, the meta-training set
is generated by exclusively drawing goal locations from the top 25\%
of the maze ($\mathcal{T}_{Train-top}$), which means that there are
10 meta-training tasks (i.e., 40 $\times$ 25\%). Subsequently, to
obtain the relevant results, we formulate two target task
distributions, one characterized by excellent alignment and the
other by weak alignment with the meta-training distribution, since
they are sampled respectively from the top 25\% portion of the maze
($\mathcal{T}_{Target-top}$) and the bottom 25\% portion of the maze
($\mathcal{T}_{Target-bottom}$). Moreover, to alleviate spurious
biases resulting from uneven density in the task distribution, we
employ density-balanced sampling tactics throughout the experimental
procedure.

Figure~\ref{fig:more_comparison_Maze}(b) and
Figure~\ref{fig:more_comparison_Maze}(c) respectively depict the
target task adaptation process with models trained under good task
alignment conditions (meta-train on $\mathcal{T}_{Train-top}$ and
meta-test on $\mathcal{T}_{Target-top}$) and bad task alignment
conditions (meta-train on $\mathcal{T}_{Train-top}$ and meta-test on
$\mathcal{T}_{Target-bottom}$). The results reveal that DCMRL indeed
can achieve superior performance with good task alignment conditions
(see Figure~\ref{fig:more_comparison_Maze}(b)). In addition, unlike
SiMPL our model trained under a misaligned meta-training task
distribution, though exhibiting initially lower performance,
eventually achieves similarly superior performance (see
Figure~\ref{fig:more_comparison_Maze}(c)). To summarize, DCMRL
exhibits strong generalization, achieving high performance with
minimal meta-training tasks and demonstrating robustness to
variations in the quality of task alignment.

\section{Conclusion}

We propose DCMRL, an offline meta-RL framework, which can acquire
more generalizable prior experience to achieve effective adaptation
to unseen target tasks. Specially, we utilize both task contexts and
skills as prior experience and use Gaussian distributions for their
representations. We extract the skills from offline datasets, and
perform exploration and learning of the continuous latent spaces of
task contexts and skills with meta-training tasks. In addition, we
propose GQ-VAE, which clusters the Gaussian distributions of task
contexts and skills in their respective continuous latent spaces and
decouples the exploration and learning processes of task contexts
and skills, enhancing their generalization. These cluster centers
which serve as representative and discrete distributions of task
context and skill are respectively stored in task context codebook
and skill codebook. Moreover, we sample positive samples, negative
samples and anchor samples through a specific sampling strategy, and
contrastively restrict the task contexts, leading to more
appropriate representations of task contexts. Experiments on
challenging continuous control navigation and manipulation tasks
that are long-horizon and sparse-reward demonstrate that DCMRL
outperforms the prior methods in meta-RL.

\appendix

\begin{center}
{\Large\textbf{Appendix}}
\end{center}

\section{Preliminaries}

\subsection{Context-based Offline Meta-RL}

Context-based offline meta-RL is a method to tackle the challenges
posed by partially observable Markov decision process (POMDP), in
which the environment states are partially observable and the agent
can only estimate them through the information it observes. A task
refers to a learning problem that contains a goal within an
environment and the task context refers to the statistics extracted
from the past experience of a task. Assuming the task $\mathcal{M}$
as the unobservable part of the states, the agent collects task
information and makes decisions with the history trajectory: $a_t
\sim \pi(a|\tau_{0:t-1},s_t)$, where $\tau_{0:t-1} = \{s_0, a_0,
r_0, \dots, s_{t-1}, a_{t-1}, r_{t-1}\}$.

One successful method for off-policy meta-RL process by building
task embeddings is probabilistic embeddings for actor-critic RL
(PEARL) \cite{DBLP:conf/icml/RakellyZFLQ19}, an online off-policy
meta-RL method which treats task context $c$ as a vector, and a
complete adaptation procedure consists of sampling $c$ from a
distribution $z \sim q_{\psi_c}(c|h)$, where the distribution
$q_{\psi_c}$ is generated by an encoder with parameters $\psi_c$ and
$h$ is the history trajectory data from meta-training tasks. This
encoder is a neural network that inputs history trajectory data
composed of a series of tuples $tuple_i = \{s_i, a_i, r_i, s'_i\}$
as elements and outputs multivariate Gaussian distribution with mean
and variance as task context. Additionally, the policy for action
decision $\pi(a|s,c)$ is conditioned on the task context $c$ which
is sampled from task context distribution by concatenating $c$ to
state $s$.

\subsection{Skill-based Offline Meta-RL}

Skill-based RL methods normally include two basic stages: skill
discovery and skill learning. In addition, a skill refers to a
sequence of actions with a fixed horizon. Some existing skill
discovery methods sample skill $z$ directly from a distribution over
skill vectors $p(z)$, and utilize this sampled skill as an input to
the skill-conditioned policy. Some other methods treat the skill
discovery stage as training model $p(z|s)$ and $p(s|z)$ based on a
distribution over states indirectly. Moreover, previous skill
learning methods belong to unsupervised skill learning and they
utilize information-theoretic approaches with interaction data,
while other methods carry out skill learning by latent variable
inference with unlabeled demonstration data.

Skill-prior RL (SPiRL) \cite{DBLP:conf/corl/PertschLL20} proposes an
effective way to compose reusable skills as short-horizon behaviors
to solve unseen long-horizon target tasks. It accelerates the
adaptation process of unseen target tasks by making use of offline
datasets towards specific tasks to extract, integrate reusable
skills and transfer them to unseen target tasks. Moreover, it mainly
includes two modules with a task-agnostic dataset, a low-level
skill-based policy $\pi(a_t|s_t, z)$ for decoding a given latent
skill into a sequence of short-horizon behaviors and a high-level
policy $\pi(Z|s)$ for guiding exploration in skill space. The target
task RL extends a classical off-policy reinforcement learning method
called soft actor-critic (SAC) \cite{DBLP:conf/icml/HaarnojaZAL18},
which guides the high-level skill policy with the learned skill
prior:
\begin{equation} \small
  \mathop{\max}_\pi\sum_t\mathbb{E}_{(s_t,z_t)\sim\rho_\pi}[r(s_t, z_t) - \nu\mathcal{D}_{KL}(\pi(Z|s_t), p(Z|s_t))],
\end{equation}
where $\mathcal{D}_{KL}$ denotes the Kullback-Leibler divergence
between current high-level policy and skill prior, and $\nu$ is a
weighting coefficient.

\section{Experimental Environments}

\begin{figure*}[t]
  \centering
  \includegraphics[width=0.86\textwidth]{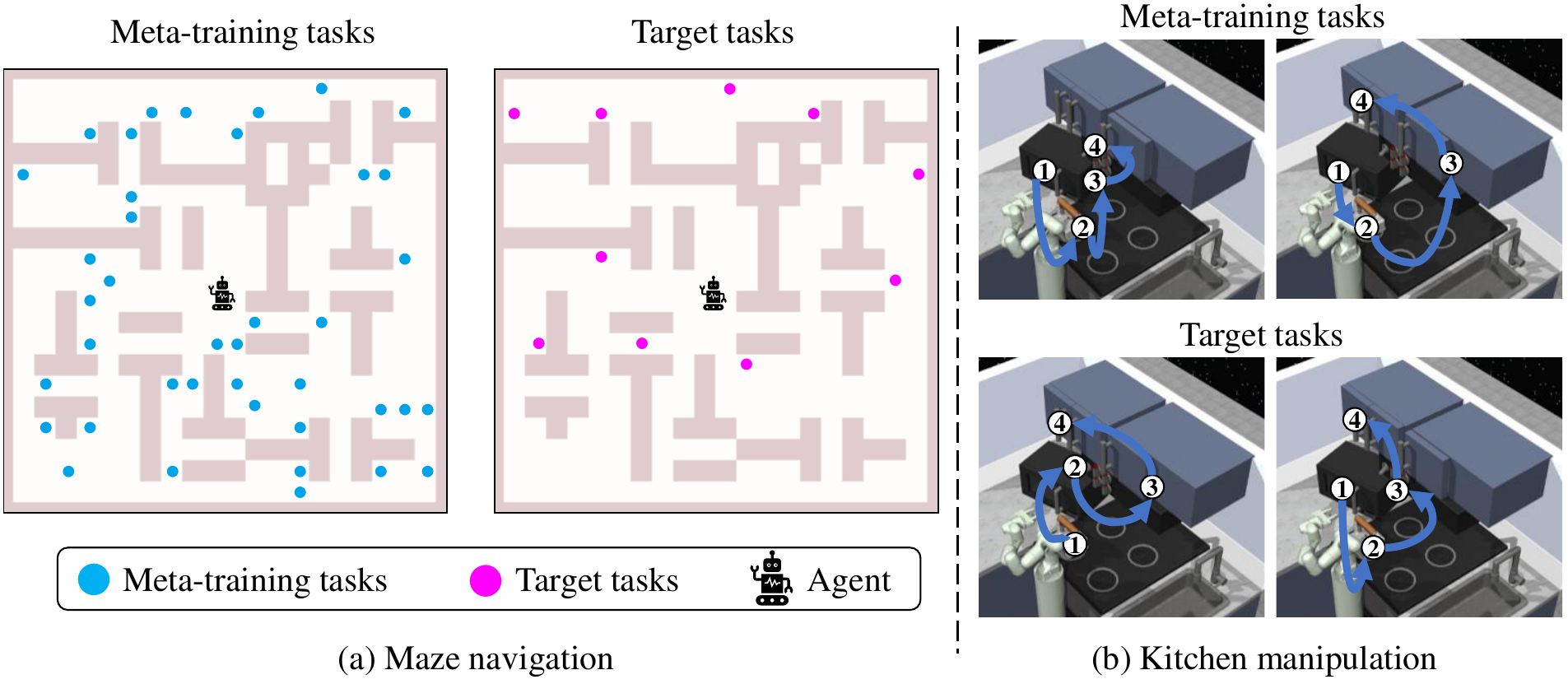}
  \caption{\textbf{Environment.} We evaluate DCMRL in two environments which
are long-horizon and spare-reward. (a) \textbf{Maze navigation}: The
agent is tasked with navigating a maze from a start point at the
center to an unknown end point where a binary reward is given upon
successful arrival within a specified step limit. (b)
\textbf{Kitchen manipulation}: The 7-DoF agent is tasked with
executing an unknown sequence of four subtasks, which are selected
from a larger set. Sparse rewards are given upon successful
completion of each subtask, with up to four sparse rewards for
executing the entire sequence in the correct order.}
  \label{fig:environment sample}
\end{figure*}

An introduction to the experimental environments, and the settings
of their meta-training and target tasks are shown in
Figure~\ref{fig:environment sample}.

\subsection{Maze Navigation}

\paragraph{Environment.}

This 2D maze navigation environment, derived from
\citet{DBLP:journals/corr/abs-2004-07219}, demands the execution of
long-horizon control strategies spanning hundreds of time steps,
while providing the agent with only sparse reward feedback upon
achieving the end goal. The state space of the agent consists of its
position and velocity coordinates in a 2D plane, and the agent's
actions are performed through the application of continuous velocity
commands in the same planar domain.

\paragraph{Offline Dataset \& Meta-training/Target Tasks.}

We follow \citet{DBLP:journals/corr/abs-2004-07219} and construct an
offline dataset devoid of task specification by randomly sampling
pairs of start and goal positions in the maze domain and utilizing a
planning algorithm to generate traversable trajectories from the
start to the end point with no assignable reward or task labels
associated. To construct both the meta-training tasks and target
tasks, the agent's initial state is uniformly fixed at the center of
the maze, followed by the selection of 40 distinct goal locations
for meta-training tasks and an additional 10 goal locations for the
set of target tasks. Moreover, sparse reward setting is used for all
meta-training and target tasks.

\subsection{Kitchen Manipulation}

\paragraph{Environment.}

The Franka kitchen environment introduced by
\citet{DBLP:conf/corl/0004KLLH19} is an environment that involves
controlling a 7-DoF robotic arm using continuous joint velocity
commands in order to execute a sequence of manipulation tasks such
as activating the stove or opening the microwave. In order to
complete a successful episode, the agent must execute a series of
subtasks, with the total length of each episode spanning 300 to 500
time steps, and is awarded sparse rewards only upon successful
completion of each subtask.

\paragraph{Offline Dataset \& Meta-training/Target Tasks.}

We follow \citet{DBLP:conf/corl/0004KLLH19} and construct an offline
dataset of 600 teleoperated manipulation sequences with human
guidance to undertake offline pre-training. Every trajectory was
designed to allow the robot to execute a sequence of four subtasks.
Based on this data, 23 different meta-training tasks and 10 specific
target tasks are identified and defined, each requiring the
consecutive execution of four subtasks. Furthermore, specific
combinations of subtasks constructed in different orders will result
in different tasks.

\section{Baselines}

\begin{compactitem}
\item SiMPL \cite{DBLP:conf/iclr/NamSPHL22}. SiMPL is a skill-based
meta-reinforcement learning approach focusing on leveraging prior
experience from large offline datasets in the absence of additional
reward or task annotations. It extracts reusable skills and skill
priors from offline data, and utilizes them in the meta-training
phase to facilitate the learning of novel, unseen target tasks. More
specifically, it particularly focuses on continuous control problems
with sparse rewards.
\item SAC \cite{DBLP:conf/icml/HaarnojaZAL18}. SAC constitutes one of the
most widely applied and advanced deep reinforcement learning
algorithms and is capable of learning how to solve a desired task
completely from scratch, with no reliance upon either offline
datasets or meta-training tasks.
\item SPiRL \cite{DBLP:conf/corl/PertschLL20}. SPiRL is a transfer
learning approach designed to harness the potential of offline data
by enhancing the acquired skills and skill priors from such
datasets. While it does avail of previously-learned skills from
offline datasets, it does not involve the utilization of
meta-training tasks. This approach hence aims to ascertain the
effectiveness of leveraging meta-training tasks for further
improvement.
\item PEARL \cite{DBLP:conf/icml/RakellyZFLQ19}. PEARL is a highly
advanced and state-of-the-art off-policy algorithm utilized in
meta-RL. It learns a policy that can rapidly adapt to unseen
prospective tasks in a given context. It mainly learns from
meta-training tasks without extending to offline datasets, aiming to
explore potential benefits accrued by integration of previously
learned skills into meta-RL techniques.
\item PEARL-ft \cite{DBLP:conf/iclr/NamSPHL22}. PEARL-ft reflects the
evaluation of an adaptive PEARL model, which is subsequently
fine-tuned on a specific target task employing the benefits of the
SAC algorithm to further augment effectiveness.
\item Multi-task RL (MTRL) \cite{DBLP:conf/nips/TehBCQKHHP17}. MTRL is a
highly effective multi-task RL benchmark that learns by distilling
individual methods optimized for each task attribute into a single
shared policy with the utilization of both meta-training tasks and
offline datasets in training.
\end{compactitem}

\section{Experimental Details}

Our experiments are conducted on a machine with NVIDIA GeForce RTX
3090 and implemented by PyTorch. DCMRL utilizes the Adam optimizer
\cite{DBLP:journals/corr/KingmaB14} with a learning rate of $3e-4$,
$\beta_1=0.9$ and $\beta_2=0.999$ to update all the networks in
meta-training and meta-testing phases.

In the meta-training phase, we set the number of context codes and
skill codes to 16 and update GQ-VAE for task contexts and skills
through Eq. (7) and Eq. (9) respectively. The weight coefficients
$\mu$, $\eta$ and $\iota$ in Eq. (5), Eq. (7) and Eq. (9) are set to
0.25. The replay buffer size is set to 20000 in the maze navigation
environment and 3000 in the kitchen manipulation environment. The
weight of triplet loss is set to $1e-4$ for the maze navigation
environment and $1e-1$ for the kitchen manipulation environment.
DCMRL updates skill encoder and critics to minimize the mean squared
error (MSE) loss between Q-value prediction and target Q-value.
Moreover, DCMRL updates all networks with the average gradients of
20 randomly sampled meta-training tasks. Each batch of gradients is
computed from 1024 transitions for maze navigation and 256
transitions for kitchen manipulation. We train the DCMRL model for
1000 episodes for the maze navigation environment with 10, 20 and 40
meta-training tasks, and 500 episodes for the kitchen manipulation
environment. DCMRL involves the application of distinct
regularization coefficients which are contingent on the size of the
conditioning transitions. We set the coefficient of target KL
divergence to 0.1 for the 4 transitions and 0.4 for the 8192
transitions in the maze navigation environment. Meanwhile, we set
the target KL divergence to 0.4 for the 1024 transitions and the
fixed KL coefficient to 0.3 for the 2 transitions in the kitchen
manipulation environment.

In the meta-testing phase, the skill policy and GQ-VAE of skills are
still in a state of training. We set the coefficient of target KL
divergence of skill policy to 0.1 in the maze navigation environment
and 2 in the kitchen manipulation environment.

\section{Ablation Studies}

\begin{figure}[t]
  \centering
  \includegraphics[width=0.4\textwidth]{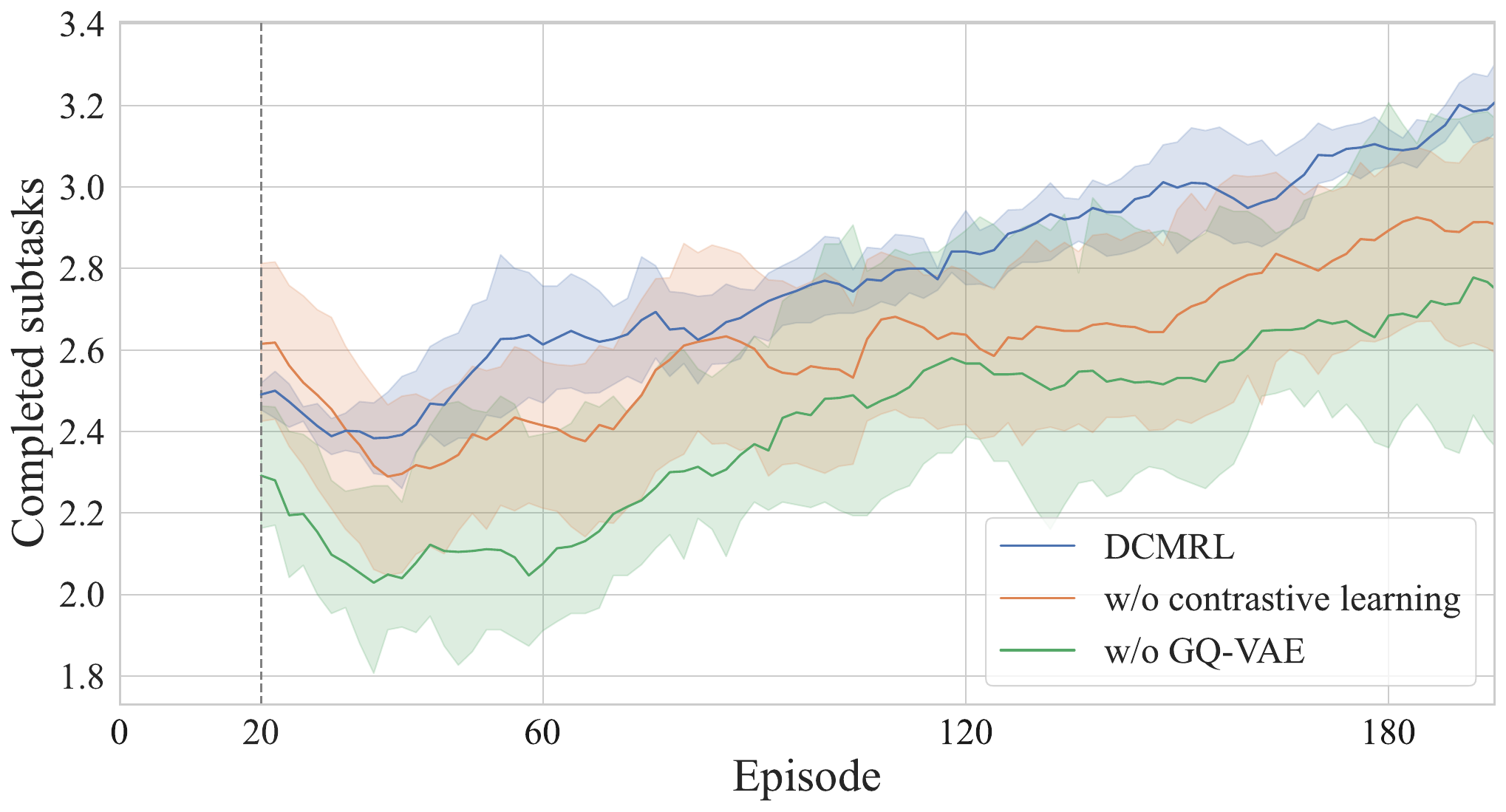}
  \caption{\textbf{Ablation experiments on modules.} We conduct ablation
experiments in the kitchen manipulation environment with two
different versions based on complete DCMRL. The version named w/o
contrastive learning is a variation that removes the contrastive
learning module, while the version named w/o GQ-VAE is a variation
that removes the GQ-VAE module.}
  \label{fig:ablation module}
\end{figure}

\begin{figure*}[t]
\centering
\includegraphics[width=0.4\textwidth]{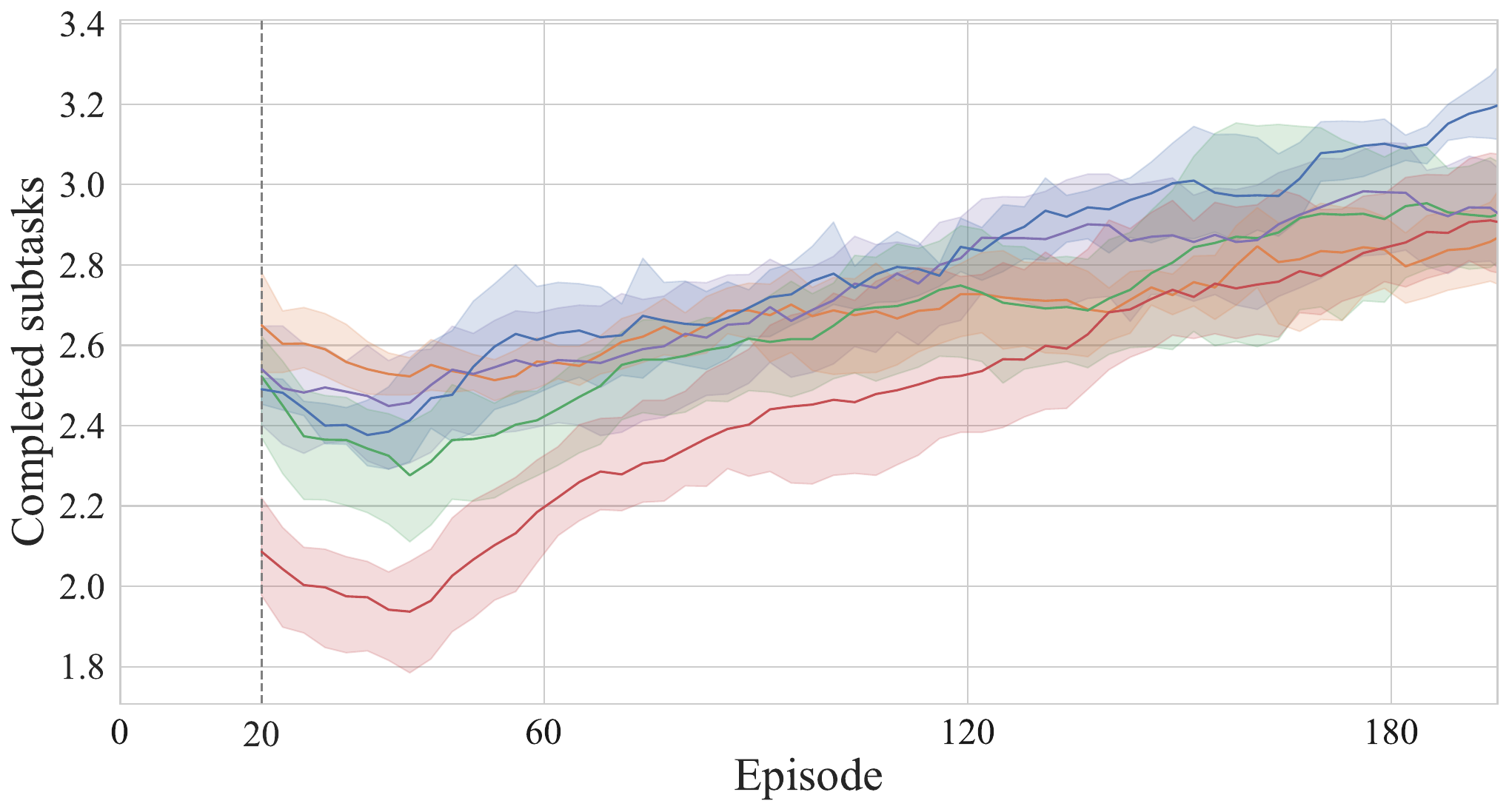}
\includegraphics[width=0.6\textwidth]{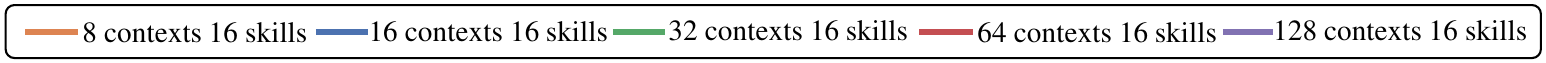}
\caption{\textbf{Ablation experiments on GQ-VAE codes for task
contexts.} With the number of codes in skill GQ-VAE fixed at 16, we
conduct separate experiments in the kitchen manipulation environment
for task context GQ-VAE under various numbers of codes, including 8,
16, 32, 64 and 128.} \label{fig:ablation context codes}
\end{figure*}

\begin{figure*}[t]
\centering
\includegraphics[width=0.4\textwidth]{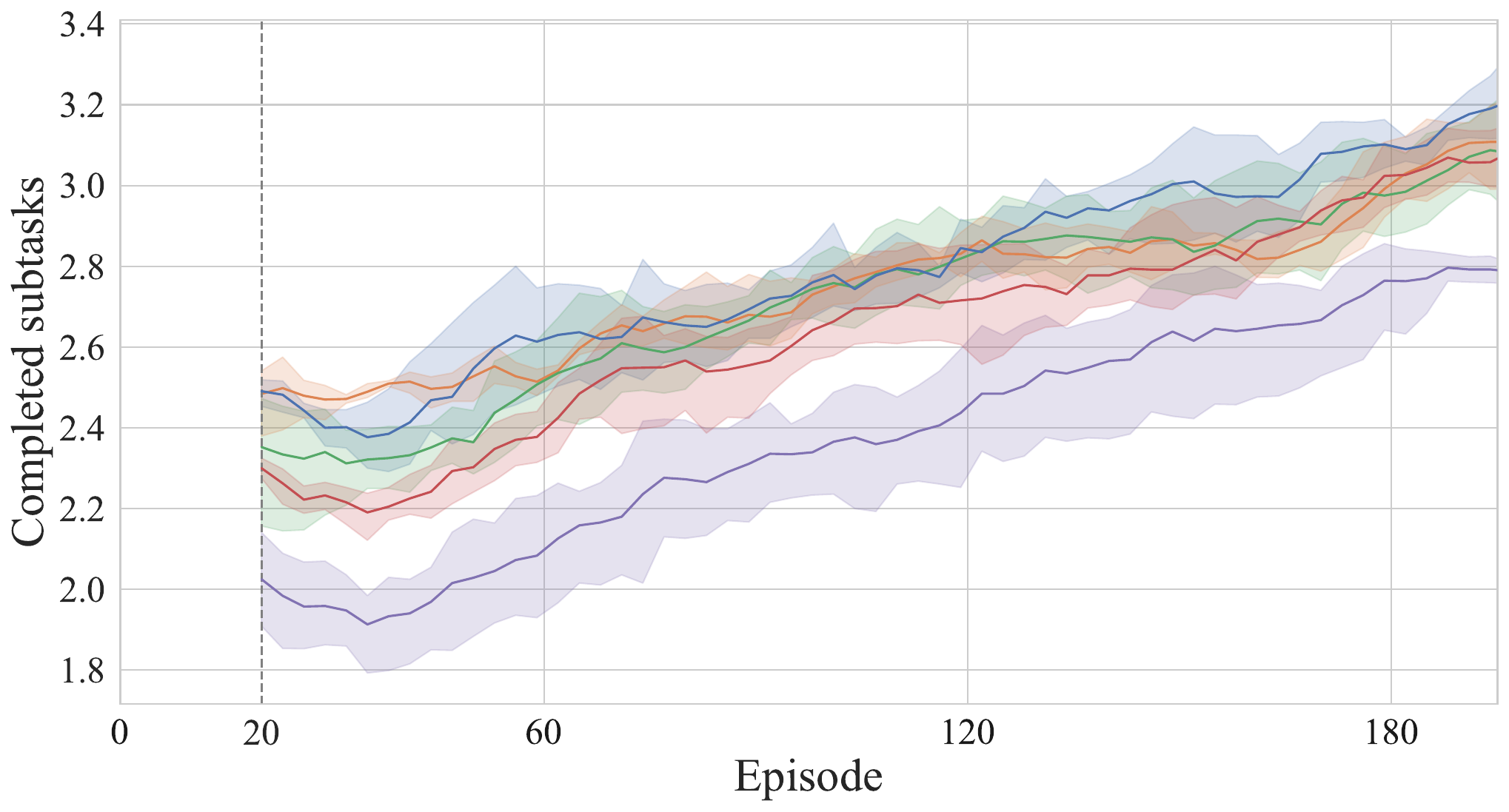}
\includegraphics[width=0.6\textwidth]{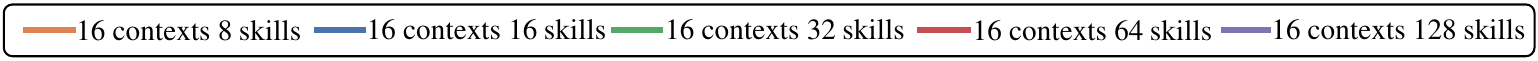}
\caption{\textbf{Ablation experiments on GQ-VAE codes for skills.}
With the number of codes in task context GQ-VAE fixed at 16, we
conduct separate experiments in the kitchen manipulation environment
for skill GQ-VAE under various numbers of codes, including 8, 16,
32, 64 and 128.} \label{fig:ablation skill codes}
\end{figure*}

\begin{figure*}[!h]
\centering
\includegraphics[width=0.4\textwidth]{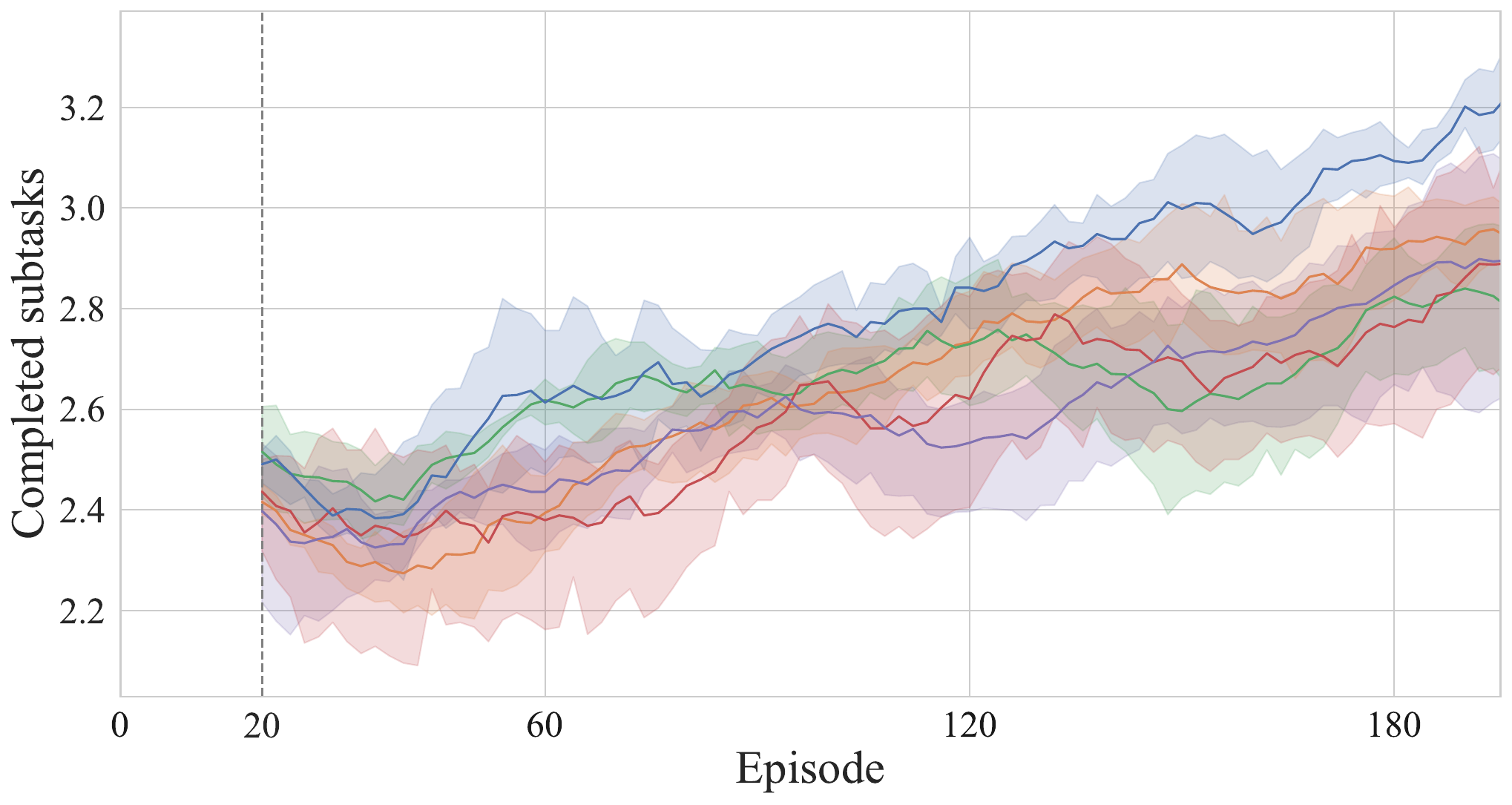}
\includegraphics[width=0.6\textwidth]{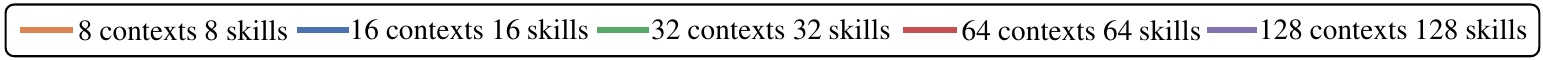}
\caption{\textbf{Ablation experiments on GQ-VAE code combinations
for contexts and skills.} We conduct experiments that simultaneously
vary the numbers of codes for both task context GQ-VAE and skill
GQ-VAE with the same number of codes in both modules. We have tested
a range of values for the number of codes, including 8, 16, 32, 64
and 128.} \label{fig:ablation both codes}
\end{figure*}

In this section, we conduct our ablation experiments in the kitchen
manipulation environment to gain a more comprehensive understanding
of DCMRL, with a specific focus on two key aspects. The first aspect
pertains to the modules in DCMRL, the contrastive learning module
and GQ-VAE (see Figure~\ref{fig:ablation module}), while the second
aspect focuses on the number of codes presented in the critical
GQ-VAE (see Figure~\ref{fig:ablation context codes},
Figure~\ref{fig:ablation skill codes} and Figure~\ref{fig:ablation
both codes}).

\subsubsection{Ablation Study of Proposed Modules}

We build two different variants, without contrastive learning and
without GQ-VAE, based on the complete DCMRL, with the main
difference being that the contrastive learning module and GQ-VAE
have been removed, respectively.

The results of ablation experiment on modules depicted in
Figure~\ref{fig:ablation module} demonstrate that the w/o
contrastive learning and w/o GQ-VAE variants exhibit similar
performance in the kitchen manipulation environment, although their
sample efficiency and performance are lower than that of DCMRL.
Additionally, both variants have demonstrated limitations in solving
more than three subtasks, which has been achieved by DCMRL.

\begin{figure*}[t]
\centering
\includegraphics[width=0.3\textwidth]{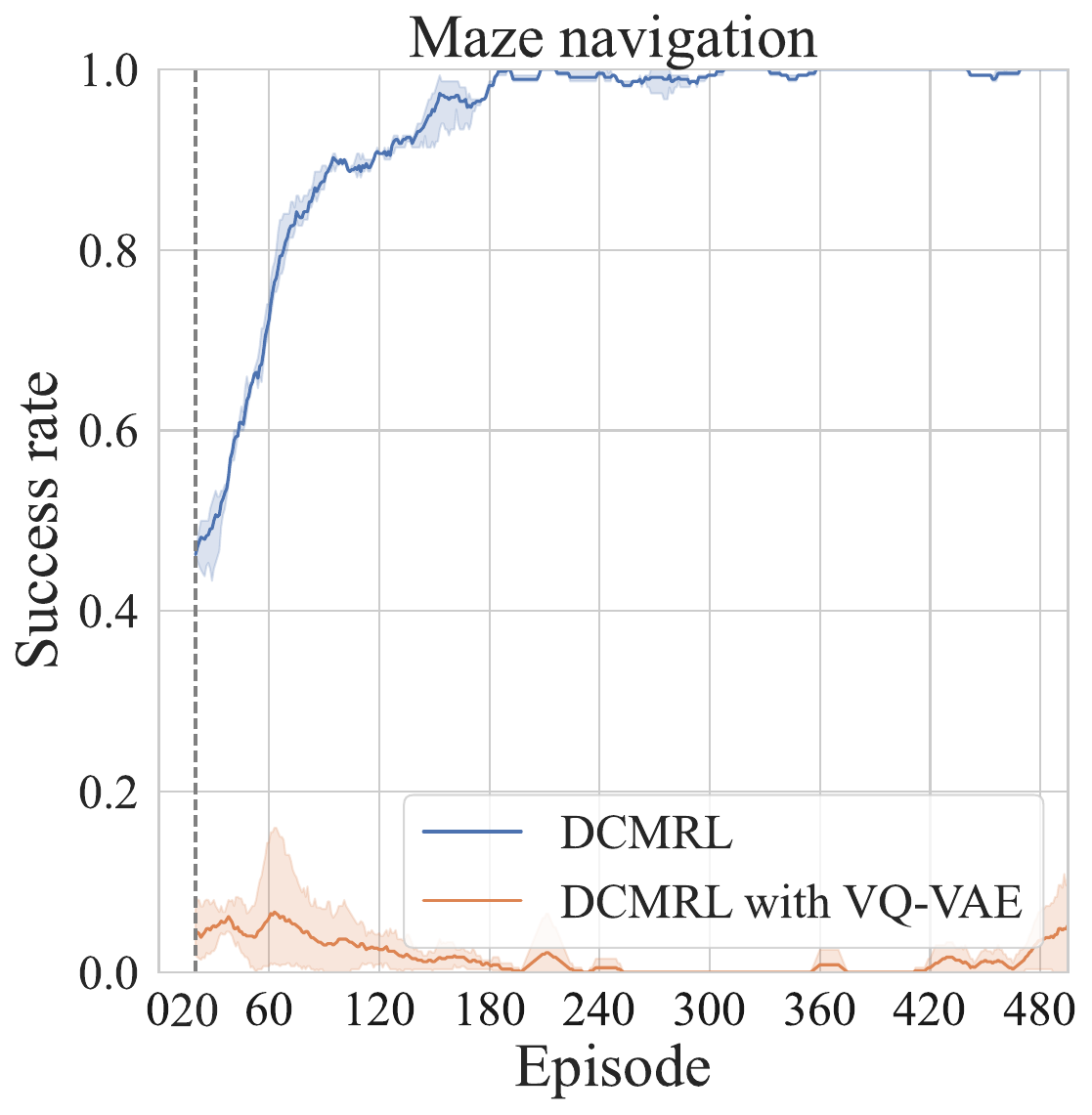}
\quad
\includegraphics[width=0.3\textwidth]{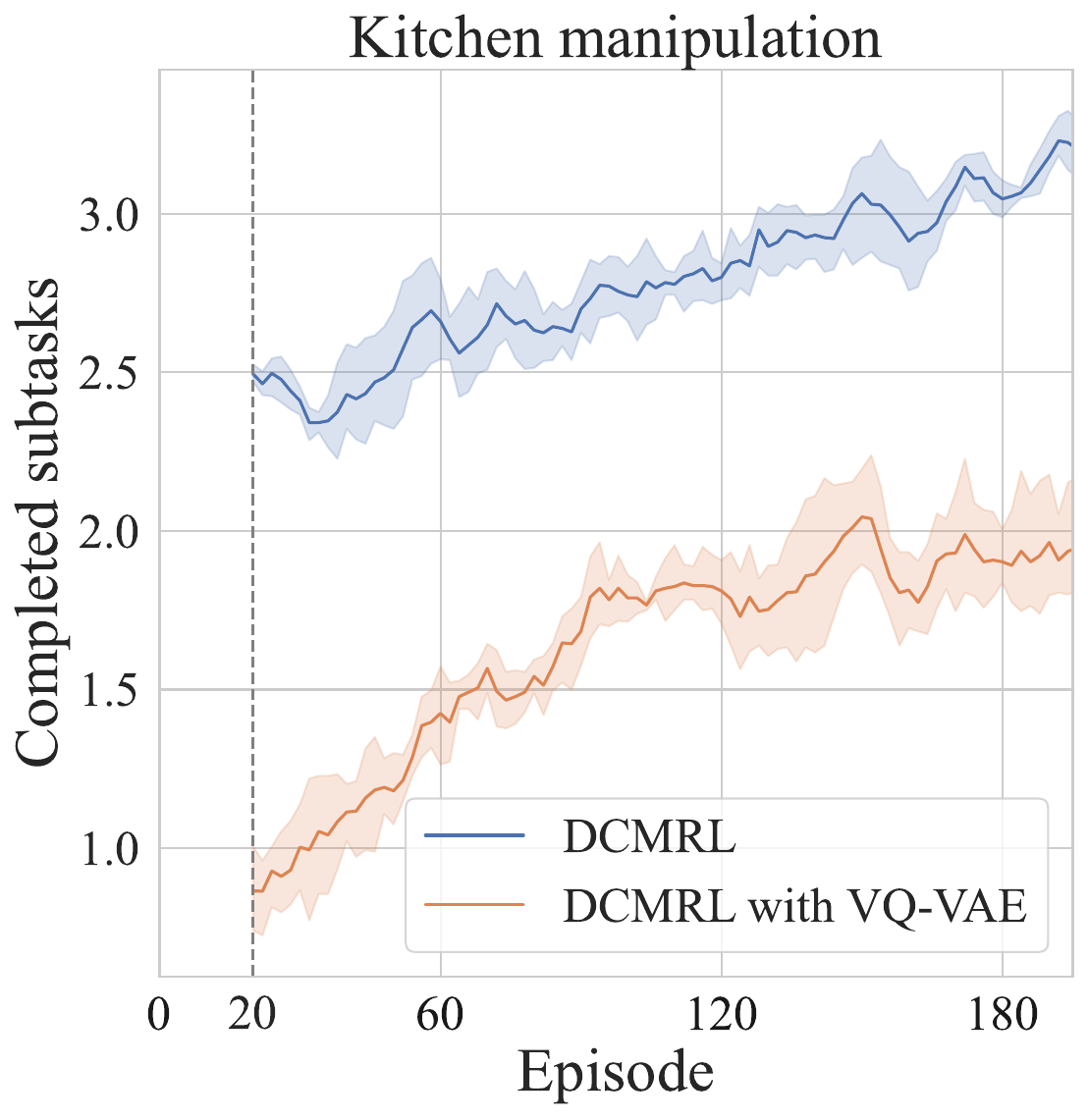}
\caption{\textbf{Sample efficiency analysis: GQ-VAE vs. VQ-VAE.} We
evaluate DCMRL and DCMRL with VQ-VAE in maze navigation and kitchen
manipulation.} \label{fig:result of VQ-VAE}
\end{figure*}

\begin{figure*}[t]
\centering
\includegraphics[width=0.7\textwidth]{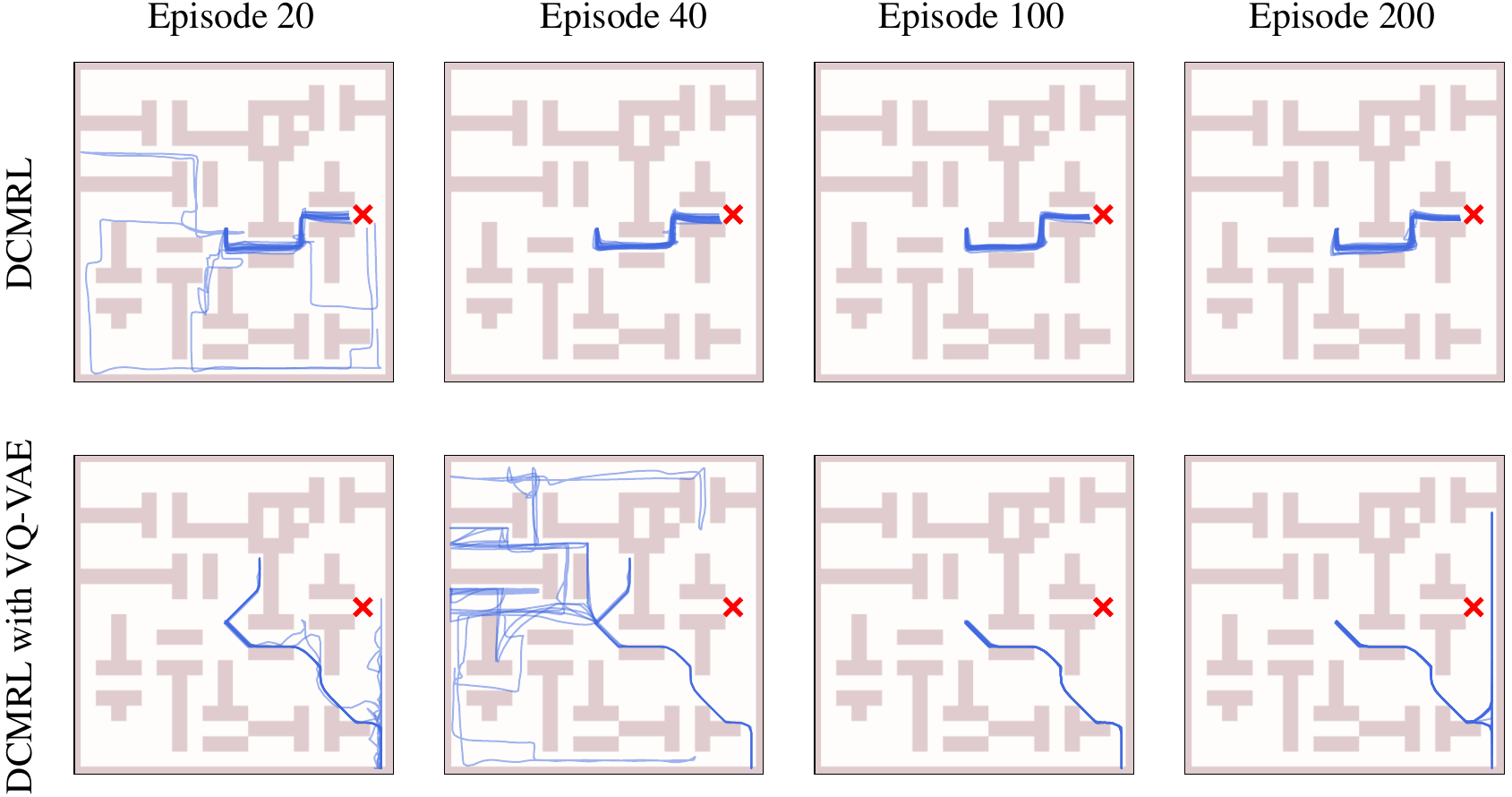}
\caption{\textbf{Visual analysis of the adaptation process.} We
present the adaptation situations of DCMRL and DCMRL with VQ-VAE in
maze navigation in episodes 20, 40, 100 and 200.}
\label{fig:visualization VQ-VAE}
\end{figure*}

\subsubsection{Effect of Code Number in GQ-VAE}

The GQ-VAE module is a key component of DCMRL, and the number of
codes in this module is an important hyperparameter. We utilize
GQ-VAE in both the task contexts and skills processing domains,
which entail hyperparameters related to the number of codes for task
contexts and skills. Moreover, we test single hyperparameter
variations as well as combinations variations of these two
hyperparameters, as shown in Figure~\ref{fig:ablation context
codes}, Figure~\ref{fig:ablation skill codes} and
Figure~\ref{fig:ablation both codes}. The results of these
experiments indicate that, the number of codes is a crucial
hyperparameter for GQ-VAE and the numbers of codes for task contexts
and skills play a similar role. When the numbers of codes in both
task context GQ-VAE and skill GQ-VAE are relatively small, as in
cases 8 contexts 8 skills (8 codes for task contexts and 8 codes for
skills), 16 contexts 8 skills and 16 contexts 16 skills, the sample
efficiency and performance is optimal. It suggests that the number
of codes like this can achieve good clustering results for the
continuous latent spaces of task contexts and skills. As the number
of codes for task contexts increases, however, as in cases 32
contexts 16 skills and 64 contexts 16 skills, the ability of DCMRL
to abstract the continuous latent space of task contexts has a
slight decrease. Meanwhile, the phenomenon also occurs when the
number of codes for skills increases, as in cases 16 contexts 32
skills and 16 contexts 64 skills, which means that the ability of
DCMRL to abstract the continuous latent space of skills also has a
comparable sample efficiency and performance. When the numbers of
codes for task contexts and skills further increase simultaneously,
the clustering ability of DCMRL has not significantly decreased, and
sample efficiency and performance still remain at a relatively
stable level, as observed in cases 64 contexts 64 skills and 128
contexts 128 skills. In summary, DCMRL can achieve excellent sample
efficiency and performance under specific hyperparameter settings,
while its sample efficiency and performance are relatively stable
under other hyperparameter settings.

\section{Comparison with VQ-VAE}

In this section, we offer a comparison of DCMRL and DCMRL with
vector quantization variational autoencoder (VQ-VAE). We report both
quantitative performance and qualitative adaptation results,
presented in Figure~\ref{fig:result of VQ-VAE} and
Figure~\ref{fig:visualization VQ-VAE} respectively.

The results in Figure~\ref{fig:result of VQ-VAE} show that DCMRL
enables better performance and higher sample efficiency in adapting
to unseen target tasks relative to DCMRL with VQ-VAE. With VQ-VAE,
DCMRL fails to adapt to unseen target tasks in the maze navigation
environment and attains poor performance in the kitchen manipulation
environment. This is because Gaussian distribution is a better form
of representing skills and task contexts compared with vector, and
our method achieves effective adaptation to unseen target tasks in
this form.

The visual representations depicted in Figure~\ref{fig:visualization
VQ-VAE} demonstrate that DCMRL with VQ-VAE completely fails to
achieve effective adaptation in the maze navigation environment.
Meanwhile, DCMRL already shows effective adaptation in the 40th
episode, but consistently poor adaptation in the case of using
VQ-VAE.

\bibliography{aaai24}

\end{document}